\documentclass{article}
\pdfoutput=1

\usepackage[final,nonatbib]{neurips_2022}

\usepackage[utf8]{inputenc} % allow utf-8 input
\usepackage[T1]{fontenc}    % use 8-bit T1 fonts
\usepackage{hyperref}       % hyperlinks
\usepackage{url}            % simple URL typesetting
\usepackage{booktabs}       % professional-quality tables
\usepackage{amsfonts}       % blackboard math symbols
\usepackage{nicefrac}       % compact symbols for 1/2, etc.
\usepackage{microtype}      % microtypography
\usepackage{xcolor}         % colors
\usepackage{amsmath}
\usepackage[pdftex]{graphicx}

\usepackage{multirow}
\usepackage{colortbl}
\usepackage{enumitem}
\usepackage{wrapfig}
\usepackage{algorithm}
\usepackage{algpseudocode}
\usepackage[figuresright]{rotating}
\usepackage{makecell}

\newcommand{\ie}{{\it i.e.}}
\newcommand{\eg}{{\it e.g.}}

\newcommand{\bd}[4]{\textbf{#1#2#3#4}}

\definecolor{grey1}{RGB}{230, 230, 230}
\definecolor{my_yellow}{RGB}{204, 167, 68}
\definecolor{my_white}{RGB}{193, 141, 158}

\title{Feature-Proxy Transformer for Few-Shot Segmentation}

\author{%
  Jian-Wei Zhang$^1$\thanks{Work done during an internship at Baidu Research.},$\quad$ Yifan Sun$^2$, $\quad$ Yi Yang$^3$, $\quad$ Wei Chen$^1$\thanks{Corresponding author.} \\
  $^1$ State Key Lab of CAD\&CG, Zhejiang University, Hangzhou, China \\
  $^2$ Baidu Research \\
  $^3$ CCAI, College of Computer Science and Technology, Zhejiang University \\
  \texttt{\{zjw.cs,yangyics,chenvis\}@zju.edu.cn, sunyf15@tsinghua.org.cn} \\
}

\begin{document}

\maketitle

\begin{abstract}
Few-shot segmentation~(FSS) aims at performing semantic segmentation on novel classes given a few annotated support samples. 
With a rethink of recent advances, we find that the current FSS framework has deviated far from the supervised segmentation framework: Given the deep features, FSS methods typically use an intricate decoder to perform sophisticated pixel-wise matching, while the supervised segmentation methods use a simple linear classification head. 
Due to the intricacy of the decoder and its matching pipeline, it is not easy to follow such an FSS framework. 
This paper revives the straightforward framework of ``feature extractor $+$ linear classification head'' and proposes a novel Feature-Proxy Transformer (FPTrans) method, in which the ``proxy'' is the vector representing a semantic class in the linear classification head. 
FPTrans has two keypoints for learning discriminative features and representative proxies: 
1) To better utilize the limited support samples, the feature extractor makes the query interact with the support features from bottom to top layers using a novel prompting strategy. 
2) FPTrans uses multiple local background proxies (instead of a single one) because the background is not homogeneous and may contain some novel foreground regions. 
These two keypoints are easily integrated into the vision transformer backbone with the prompting mechanism in the transformer. Given the learned features and proxies, FPTrans directly compares their cosine similarity for segmentation. Although the framework is straightforward, we show that FPTrans achieves competitive FSS accuracy on par with state-of-the-art decoder-based methods. \footnote[1]{Code is available at \url{https://github.com/Jarvis73/FPTrans}.}

\end{abstract}

\section{Introduction}\label{intro}

Few-shot learning is of significant value for semantic segmentation. It is because the semantic segmentation task requires pixel-wise annotation, which is notoriously cumbersome and expensive~\cite{pascal-voc-2012, linMicrosoft2014,liFSS10002020}. Therefore, learning from very few samples for semantic segmentation has attracted significant research interest, yielding a popular topic, \emph{i.e.}, few-shot semantic segmentation (FSS). Formally, FSS aims at performing semantic segmentation on novel classes given only a few (\eg, one or five) densely-annotated samples (called \emph{support} images)~\cite{caellesOneShot2017}.

\begin{figure}[t]
    \centering
    \includegraphics[width=\linewidth]{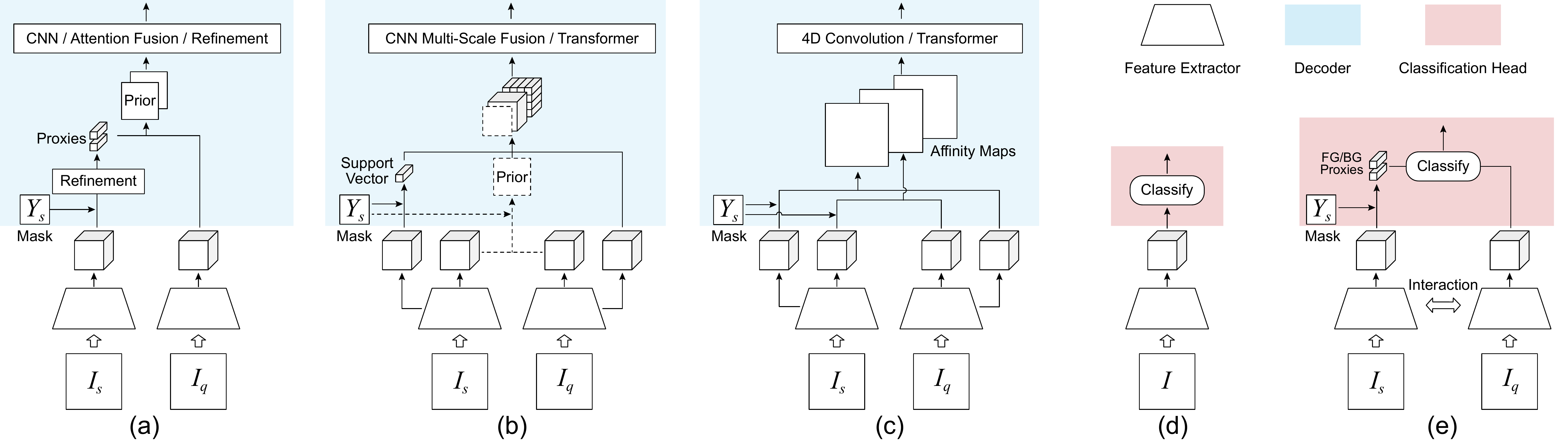}
    \caption{Comparison between the ``feature extractor + intricate decoder'' (a, b and c) and the plain``feature extractor + linear classification head'' framework (d and e). 
    \textbf{(a)} The decoder refines the prior maps (raw segmentation) for final prediction~(\eg, FWB~\cite{Nguyen_2019_ICCV}, CWT~\cite{luSimpler2021}, SCL~\cite{zhangSelfGuided2021}). \textbf{(b)} The decoder concatenates the support prototype and the query features and then further feeds them into the CNN or transformer~(\eg, PFENet~\cite{tianPrior2020}, ASGNet~\cite{liAdaptive2021}, CyCTR~\cite{zhang2021few}). \textbf{(c)} The decoder conducts pixel-to-pixel matching from query to support and then applies 4D convolution or transformer to discover the patterns within the matching score maps~(\eg, HSNet~\cite{minHypercorrelation2021}, VAT~\cite{hongCost2021}). \textbf{(d)} The plain framework of ``feature extractor + linear classification head'' in supervised segmentation. \textbf{(e)} The proposed FPTrans revives the plain framework and makes only one necessary modification (\emph{i.e.}, the proxies are extracted from the support images on the fly).}
    \label{fig:intro}
    \vspace{-5pt}
\end{figure}

With a rethink of recent advances in FSS, we find current FSS methods usually require an intricate decoder, deviating far from the plain supervised segmentation framework. More concretely, state-of-the-art FSS methods adopt the ``feature extractor $+$ (intricate) decoder'' framework (Fig.~\ref{fig:intro} (a), (b) and (c)), while the supervised segmentation methods usually adopt the ``feature extractor $+$ (simple) linear classification head'' framework (Fig.~\ref{fig:intro} (d)). In FSS frameworks, the decoders perform sophisticated matching and can be summarized into three types (Fig.~\ref{fig:intro} (a), (b), and (c)), as detailed in the related works in Section\ref{sec: relatedwork_FFS}. 
Arguably, the intricacy of the decoder and its sophisticated matching pipeline makes the FSS framework hard to follow. Under this background, we think it is valuable to explore a relatively straightforward FSS framework. 
%In Fig.~\ref{fig:intro}(a), some methods (\eg, FWB~\cite{Nguyen_2019_ICCV}, CWT~\cite{luSimpler2021}, SCL~\cite{zhangSelfGuided2021}) generate prior maps based on query and support features. They further use CNN or transformer to refine the prior maps (which may be viewed as raw segmentation) into the final segmentation output. In Fig.~\ref{fig:intro}(b), some methods (\eg, PFENet~\cite{tianPrior2020}, ASGNet~\cite{liAdaptive2021}, CyCTR~\cite{zhang2021few}) concatenate the support prototype vector  and the query features and then feed the concatenated feature maps into a sub-sequential CNN or transformer for prediction. In Fig.~\ref{fig:intro}(c), some methods (\eg, HSNet~\cite{minHypercorrelation2021}, VAT~\cite{hongCost2021}) calculate the pixel-to-pixel matching scores between support and query features to derive affinity maps.They further apply 4D convolution or transformer to discover the latent patterns within the affinity maps. 
% Arguably, the intricacy of the decoder and its sophisticated matching pipeline makes the FSS framework hard to follow. Under this background, we think it is valuable to explore a relatively simple and straightforward FSS framework. 

This paper revives the plain framework of ``feature extractor $+$ linear classification head'' for FSS and proposes a novel Feature-Proxy Transformer (FPTrans) method. The term ``proxy'' denotes the vector representing a foreground class or the background in the linear classification head. We note that in Fig.~\ref{fig:intro} (d), given the extracted feature maps, the supervised segmentation methods simply feed them into a linear classification head to perform pixel-wise prediction. FPTrans adapts this straightforward framework to the FSS task with only one modification: instead of fixing the already-learned proxies in the classification head (Fig.~\ref{fig:intro} (d)), FPTrans uses support feature maps and the support mask to generate the proxies on the fly. 
This modification is necessary for recognizing novel classes and is consistent with some earlier FSS methods \cite{shaban2017one,Wang_2019_ICCV}. 
%In Fig.~\ref{fig:intro} (d), the proxies in the classification head are usually trained through gradient back-propagation and fixed for testing. In contrast, in FPTrans (Fig.~\ref{fig:intro} (e)) the proxies in the classification head are generated on-the-fly based on the support feature maps and mask. This modification is necessary for recognizing novel classes in FSS and is consistent with some earlier FSS methods \cite{shaban2017one,Wang_2019_ICCV}.

To tackle two FSS challenges (\emph{i.e.}, generalization to novel classes and very few support samples) under this simple framework, FPTrans has two keypoints for learning discriminative features and representative proxies, respectively. \textbf{1)} To better utilize the limited support samples, the feature extractor makes the query interact with the support features from the bottom to top layers. Consequently, the support sample provides extra information/clues for extracting the query features and is thus beneficial. \textbf{2)} To promote generalization to novel classes, FPTrans uses multiple local background proxies instead of a single 
global proxy for representing the background. 
This design is important because, during base training, the background is not homogeneous and may contain some novel foreground classes. Consequently, it avoids confusing novel classes with the background of the base data and thus benefits generalization to novel classes. 

We implement the above two keypoints with a novel prompting strategy.
While using the prompt to condition a transformer for different tasks \cite{liPrefixTuning2021,jiaVisual2022,wangLearning2022} or different domains \cite{geDomain2022} is a common practice, our prompting strategy is significantly different and has two novel functions, \emph{i.e.}, 1) prompting for different (foreground or background) proxies and, 2) acting as the intermediate for query-support interaction. Specifically, FPTrans simultaneously prepends multiple prompts at the input layer, with one prompt for the foreground and the other prompts for the background. These prompts are fed into the transformer and finally become the foreground proxy and local background proxies, respectively. During their flow from the input layer to the output layer, the hidden states of prompts in all the hidden layers are shared by the query and support for cross-attention (query-prompt attention and support-prompt attention). Therefore, it significantly reduces the interaction complexity from $O(N^2)$ to $O(N)$ ($N$ is the number of pixels on the feature maps). Since the prompting and attention mechanism are critical for these two keypoints, we use the transformer backbone as a natural choice. 

%Considering the keypoints for learning ``feature'' and ``proxy'', as well the plain framework of ``feature extractor $+$ linear classification head'', we name our method as Feature-Proxy Transformer. 
We conduct extensive experiments and show that FPTrans achieves accuracy on par with the decoder-based FSS methods. For example, on PASCAL-5$^i$~\cite{caellesOneShot2017} with one support sample, FPTrans achieves 68.81\% mIoU, setting a new state of the art. With its simple framework and competitive accuracy, we hope FPTrans can serve as a strong baseline for FSS.

To sum up, our main contributions are summarized as follows: 
(i) We revive the plain ``feature extractor $+$ linear classification head'' framework for FSS. We hope the correspondingly-proposed method FPTrans can serve as a simple and strong FSS baseline. 
(ii) We integrate two keypoints into FPTrans, \emph{i.e.}, learning discriminative features through query-support interaction and learning representative local background proxies. These two keypoints rely on a novel prompting strategy of the transformer backbone and correspondingly tackle two FSS challenges, \emph{i.e.}, very few support samples and generalization to novel classes. 
(iii) We conduct extensive experiments to validate the effectiveness of FPTrans. Experimental results show that FPTrans with a plain framework achieves competitive accuracy, compared with state-of-the-art FSS methods with intricate decoders.

\section{Related Work}\label{sec: relatedwork}

\subsection{Recent Progress on Few-Shot Segmentation}\label{sec: relatedwork_FFS}
%In FSS frameworks, the decoders perform sophisticated matching and can be summarized into three types (Fig.~\ref{fig:intro} (a), (b) and (c)). 
%In Fig.~\ref{fig:intro}(a), some methods (\eg, FWB~\cite{Nguyen_2019_ICCV}, CWT~\cite{luSimpler2021}, SCL~\cite{zhangSelfGuided2021}) generate prior maps based on query and support features. They further use CNN or transformer to refine the prior maps (which may be viewed as raw segmentation) into the final segmentation output. In Fig.~\ref{fig:intro}(b), some methods (\eg, PFENet~\cite{tianPrior2020}, ASGNet~\cite{liAdaptive2021}, CyCTR~\cite{zhang2021few}) concatenate the support prototype vector  and the query features and then feed the concatenated feature maps into a sub-sequential CNN or transformer for prediction. In Fig.~\ref{fig:intro}(c), some methods (\eg, HSNet~\cite{minHypercorrelation2021}, VAT~\cite{hongCost2021}) calculate the pixel-to-pixel matching scores between support and query features to derive affinity maps.They further apply 4D convolution or transformer to discover the latent patterns within the affinity maps. 

%Few-shot segmentation is derived from the few-shot learning~\cite{vinyalsMatching2016,snellPrototypical2017} aims at tackling the problem of segmenting unseen classes with only a few densely annotated samples. 
%Most FSS methods adopt a ``feature extractor $+$ (intricate) decoder'' framework to extract features and perform support/query matching as shown in Fig.~\ref{fig:intro}(a),(b), and (c). 
Early FSS methods directly generate class-specific classifier weights and biases~\cite{shaban2017one,rakelly2018conditional,tianDifferentiable2019} or apply the prototype learning~\cite{snellPrototypical2017,dong2018few,Wang_2019_ICCV,Siam_2019_ICCV}. The recent state-of-the-art methods shift to the decoder-based framework and can be categorized into three types according to the decoder structure. \textbf{1)} Some recent methods\cite{Nguyen_2019_ICCV,liuCRNet2020,luSimpler2021,zhangSelfGuided2021} (Fig.~\ref{fig:intro}(a)) generate prior maps based on query features and support features. They further use CNN or transformer to refine the prior maps (which may be viewed as raw segmentation) into the final segmentation output. \textbf{2)} Some methods~\cite{zhangSGOne2020,Zhang_2019_CVPR,tianPrior2020,liAdaptive2021,zhang2021few,xieScaleAware2021,xieFewShot2021,yangPrototype2020,Zhang_2019_ICCV} (Fig.~\ref{fig:intro}(b)) concatenate the support prototype vector and the query features and then feed the concatenated feature maps into a subsequential CNN or transformer for prediction. %Considering to enhance class-wise representation, multiple prototypes ~\cite{yangPrototype2020,Zhang_2019_ICCV,liAdaptive2021,zhangSelfGuided2021,liuPartAware2020} are explored in feature space with EM algorithm, clustering, and so on. 
\textbf{3)} Some methods focus on exploring fine-grained knowledge and calculating the pixel-to-pixel matching scores between support and query features to derive affinity maps~\cite{huAttentionBased2019,wangFewShot2020a,yangBriNet2020,gairolaSimPropNet2020,zhang2021few,yangNew2019,luSimpler2021} (Fig.~\ref{fig:intro}(c)). Some further apply 4D convolution or transformer~\cite{minHypercorrelation2021,hongCost2021} to discover the latent patterns within the affinity maps. Besides, latent information is also investigated to enhance the model~\cite{pambalaSML2020,zhaoObjectnessAware2020,khandelwalUniT2021,yangMining2021,wuLearning2021,liuPartAware2020}. 

In contrast, this paper abandons the intricate decoder and revives the plain framework of ``feature extractor + linear classification head''. We show that this simple framework can also achieve promising FSS results. We note that a recent work~\cite{boudiafFewShot2021} also uses the plain framework during training. However, they rely on transductive inference during testing to mitigate the gap between the inconsistent training and testing schemes. Compared with \cite{boudiafFewShot2021}, the proposed FPTrans maintains its simplicity for testing and achieves superior FSS accuracy.

\subsection{Backbone for Few-Shot Segmentation}
Previous FSS methods usually adopt CNNs (\eg, VGG~\cite{simonyanVery2015}, ResNet~\cite{heDeep2016}) as the backbone (\ie, feature extractor) and typically fix the pretrained backbone parameters~\cite{Zhang_2019_CVPR,tianPrior2020,zhang2021few,zhangSelfGuided2021,wuLearning2021,luSimpler2021,liAdaptive2021,minHypercorrelation2021,hongCost2021}. However, fixing the backbone is prone to a side-effect, \emph{i.e.}, insufficient adaptation to the segmentation training data. Some recent methods finetune the CNN backbone along with FSS training with sophisticated techniques (\emph{e.g.}, model ensemble or transductive inference~\cite{langLearning2022,boudiafFewShot2021}) to tackle the problem of insufficient adaptation. %However, they rely on extra intricate decoders~\cite{minHypercorrelation2021,hongCost2021} or techniques~(\eg, ensemble, transductive inference)~\cite{langLearning2022,boudiafFewShot2021} to reach the desired performance. 

In contrast, this paper adopts the vision transformer~\cite{dosovitskiyImage2021,henaffDataEfficient2020,xieSegFormer2021,liuSwin2021} as the backbone because we rely on the attention mechanism and a novel prompting technique for query-support interaction. Moreover, the proposed FPTrans benefits from fine-tuning the backbone parameters without bells and whistles. 
%Transformer constructed based on attention~\cite{vaswaniAttention2017} is feasible to extend model inputs for introducing novel class knowledge, which is critical for the FSS problem. 2) With a simple proxy-based linear classification head, we can directly finetune the transformer backbone for achieving competitive FSS accuracy. 
We note that using the transformer backbone does NOT necessarily improve FSS, because the ablation studies~(in Table~\ref{tab:backbone}) show that replacing the CNN backbone with a transformer does NOT bring improvements to the decoder-based FSS methods~\cite{tianPrior2020, zhang2021few}. Therefore, we attribute the superiority of our method mainly to the two unique keypoints in FPTrans. %In other words, using the transformer backbone does NOT necessarily improves FSS.  %Therefore, using the transformer backbone is a natural choice for FPTrans with plain framework, while using the CNN backbone is 

\section{Methods}\label{methods}

\subsection{Problem Formulation}

Few-shot segmentation aims at tackling the semantic segmentation problem on novel classes under a low data regime. Specifically, FSS usually provides a training set with categories $\mathcal{C}_{train}$ and a testing set with novel categories $\mathcal{C}_{test}$ ($\mathcal{C}_{train}\cap\mathcal{C}_{test}=\emptyset$). %Given only few support images for each testing category, FSS seeks for accurate segmentation on query images. 
%where $\mathcal{C}_{train}\cap\mathcal{C}_{test}=\emptyset$ is held. 
The mainstream setting~\cite{Zhang_2019_CVPR,zhang2021few,tianPrior2020} adopts the episodic training and testing scheme: Each episode corresponds to a single class $c$ ($c\in\mathcal{C}_{train}$ during training and $c \in \mathcal{C}_{test}$ during testing), and provides a query sample $\{I_q, Y_q\}$ and $K$ support samples $\{I_s^{(k)}, Y_s^{(k)}\}_{k=1}^K$ ($I$ is the image and $Y$ is the label). The superscript $k$ will be omitted unless necessary. Within each episode, the model is expected to use $\{I_s, Y_s\}$ and $I_q$ to predict the query label. In this paper, we follow this popular episodic training and testing scheme. 

%In an episode, the corresponding class is regarded as the foreground class; Other classes are merged into the background class. 
%Given an episode, the model is desired to use $I_q,I_s$ and $Y_s$ to predict the query mask.
%, which is formulated as:
%\begin{equation}
%    \mathcal{L} = CE(F_{\theta}(I_q, I_s, Y_s), Y_q).
%\end{equation}
%$CE$ is a two-class~(foreground and background) cross entropy loss, and $F_{\theta}$ is a model parameterized by $\theta$. When performing inference, only $I_q, I_s$ and $Y_s$ are accessible, and $Y_q$ is to be predicted.

%==================================================================================
\subsection{Preliminaries on Vision Transformer and Prompt Learning}
The proposed FPTrans uses the vision transformer as its backbone and integrates a novel prompting strategy. Therefore, we first revisit the vision transformer and the prompting mechanism. 

\textbf{Vision Transformer}~\cite{dosovitskiyImage2021}(ViT) is designed for computer vision tasks based on transformer~\cite{vaswaniAttention2017} which is originally designed for sequential data~\cite{devlin2019BERT,brownLanguageModelsAre2020,Yijing2022TPGNN}. It is composed of a patch embedding module, a transformer encoder, and an MLP head. Given an RGB image as the input, ViT first reshape it into $N$ patches $\{\mathbf{a}_p\in\mathbb{R}^{3\times P\times P}\vert p=1,2,\dots,N\}$ ($P$ is the patch size) and then projects the flattened image patches into $C$-Dimensional embeddings by $\mathbf{x}_p=\text{Embed}(\mathbf{a}_p)\in\mathbb{R}^C,\; p=1,2,\dots,N$. We denote the collection of these embedding tokens as $\mathbf{X}^0 = \{\mathbf{x}_p\}_{p=1}^N \in \mathbb{R}^{N\times C}$. ViT has $L$ stacked transformer blocks, each one of which consists of a Multiheaded Self-Attention module and an MLP module (See supplementary Section A for details). Given $\mathbf{X}^0$ (the collection of embedding tokens), ViT concatenates it with a classification token $\mathbf{x}_{cls}^0 \in \mathbb{R}^C$ and then inputs them into the stacked transformer blocks,  which is formulated as:
\begin{equation}\label{eqn:vit1}
    [\mathbf{x}_{cls}^l, \mathbf{X}^l] = B_l([\mathbf{x}_{cls}^{l-1}, \mathbf{X}^{l-1}]), \qquad l=1,2,\dots,L, 
    % \mathbf{y} &= \text{MLP\_Head}(\mathbf{x}_{cls}^L),
\end{equation}
where $\mathbf{x}_{cls}^l$ and $\mathbf{X}^l$ are the outputs of the $l$-th block $B_l$, and $[\cdot,\cdot]$ is the concatenate operation. 

\textbf{Prompting} was first introduced in NLP tasks to identify different tasks by inserting a few hint words into input sentences~\cite{gao-etal-2021-making,jiangHow2020, brownLanguage2020,liuPretrain2021}. More generally, prompting techniques can efficiently condition the transformer to different tasks~\cite{liPrefixTuning2021,guPPT2022,houlsbyParameterEfficient2019,heUnified2022,wangLearning2022} or domains~\cite{geDomain2022} without changing any other parameters of the transformer. To this end, prompting techniques typically prepend some prompt tokens $\mathbf{P}^0$ to the input layer. Correspondingly, Eqn~\eqref{eqn:vit1} transforms into  $[\mathbf{x}_{cls}^l,\mathbf{X}^l,\mathbf{P}^l]=B_l([\mathbf{x}_{cls}^{l-1},\mathbf{x}^{l-1},\mathbf{P}^{l-1}])$. It can be seen that changing the prompt simultaneously changes the mapping function of the transformer, even if all the transformer blocks $B_l$ remain  unchanged. 

\textbf{Our prompting strategy} in FPTrans is significantly different from the prior prompting techniques. In contrast to the popular prompting for different tasks or different domains, FPTrans prepends multiple prompts to simultaneously activate multiple different proxies (\emph{i.e.}, the foreground and local background proxies), as well as to facilitate efficient query-support interactions. Moreover, in prior works, the prompts in the hidden layers are dependent on a single input sample. In contrast, in FPTrans, the prompts in the hidden layers are shared by the query and support images through a synchronization procedure. We will illustrate these points in the following section. 

\subsection{Feature-Proxy Transformer}\label{sec:FPTrans}

\begin{figure}
    \centering
    \includegraphics[width=\linewidth]{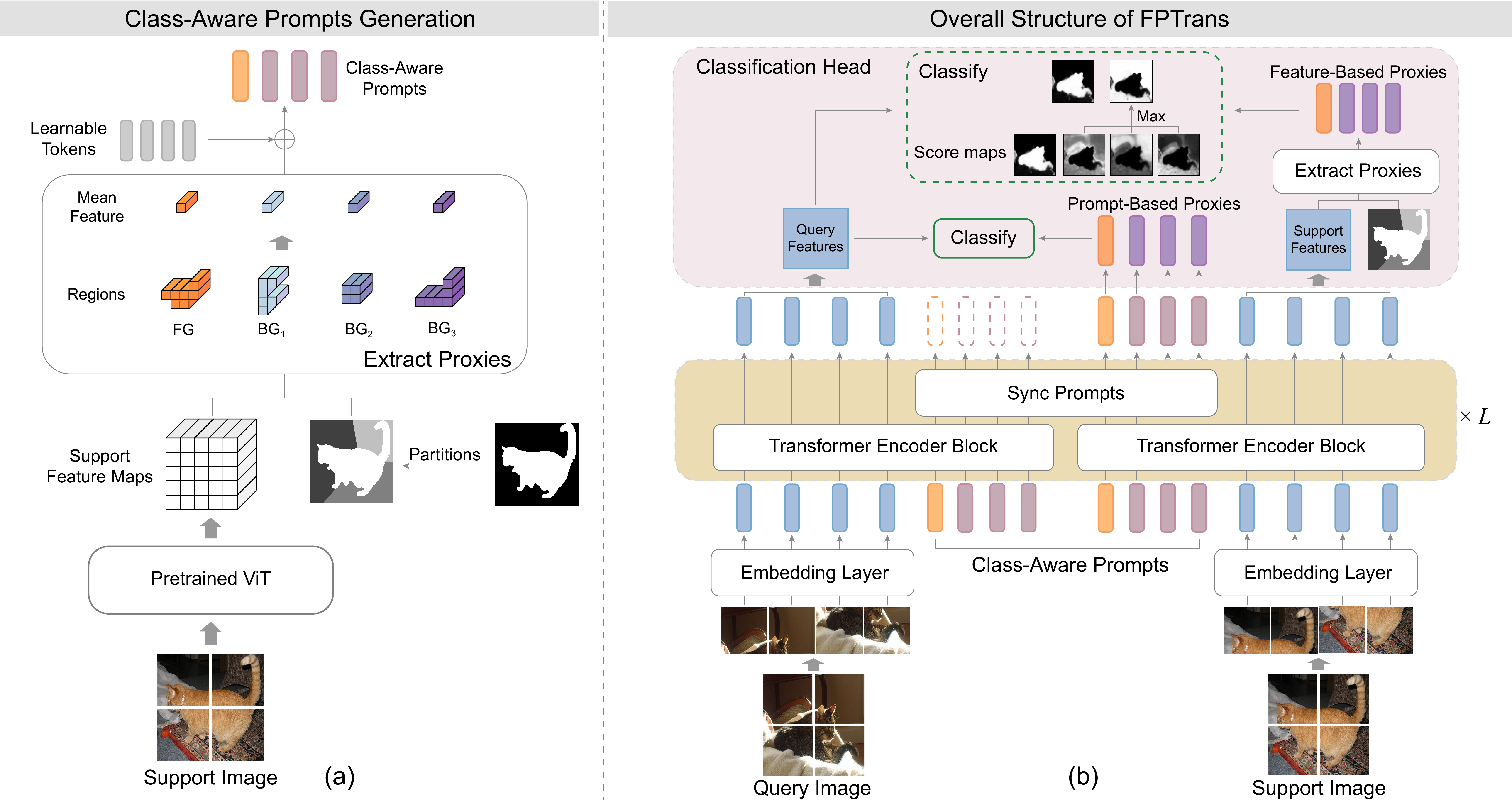}
    \caption{Overview of the proposed {Feature-Proxy Transformer}~(FPTrans). \textbf{(a)} Given a support image, we generate a foreground prompt and multiple local background prompts. Each prompt consists of multiple tokens.
    \textbf{(b)} The \textcolor{my_yellow}{feature extractor} consists of $L$ transformer blocks. It takes patch tokens from the query and support images, as well as the prompts as its input. 
    %Given both the support image and query image, FPTrans feeds the image patch tokens and the prompt tokens into the \textcolor{yellow}{feature extractor} stacked with $L$ transformer blocks.  
    After every transformer block, FPTrans synchronizes the prompt tokens from the query and support branches to facilitate efficient query-support interactions. The \textcolor{my_white}{ classification head} uses two types of proxies (the feature-based and the prompt-based proxies) for training and uses the feature-based proxies for inference. }
    %Class-aware prompts generation and multiple local background region partition. \textbf{(b)} The structure of FPTrans. Query and support features are extracted by a transformer backbone, where the class-aware prompts are used to introduced current novel class information and enable interaction between support and query with a ``Sync Prompts'' module. From the support features and the deep states of prompts, we calculate feature-based proxies and prompt-based proxies, with which a simple classification head is applied for query prediction.}
    \label{fig:methods}
\end{figure}

\subsubsection{Overview}
The proposed Feature-Proxy Transformer~(FPTrans) consists of three major steps, \emph{i.e.}, 1) prompt generation, 2) feature and proxy extraction, and 3) training the classification head and inference, as illustrated in Figure ~\ref{fig:methods}. The  prompts are extracted from the support images within a (training or testing) episode and contain a foreground prompt and multiple local background prompts (Section~\ref{sec:prompt}). Afterward, FPTrans concatenates these prompts with the image patch tokens and forwards them into the stacked transformer blocks to extract both the features and proxies (Section~\ref{sec:feature_proxy}). Different from the popular prompting technique, FPTrans shares the hidden states of the prompt tokens for the query and its support images, using a prompt synchronization operation. It facilitates efficient feature interactions between support and query features. Finally, given the extracted features and proxies, we elaborate on how to use them for training the classification head and for inference in Section~\ref{sec:train_infer}.

% To better utilize the limited support samples, we first generate a group of prompt tokens from support samples with a pretrained vision transformer. The prompts consist of a foreground prompt and multiple local background prompts, representing the current foreground class and the inhomogeneous background. Then, the prompt tokens shared between support and query, are combined with image tokens as the inputs of feature extractor. During the forward process, we synchronize the hidden states of prompt tokens between support and query hidden layers, which allows the query features to interact with support features in an efficient way. With the deep features generated by the feature extractor, we calculate a foreground proxy and multiple local background proxies and use them to classify query features, which is also applied with the output states of prompts. 

\subsubsection{Prompt Generation}\label{sec:prompt}

Prompt generation is illustrated in Fig.~\ref{fig:methods} (a). Within each episode, FPTrans uses the support image(s) to extract class-aware prompts and share them for the support and query images. The prompts are class-aware because we respectively extract prompts from the foreground and  background. %In other words, the foreground and background have their own prompts. 

To this end, we first use a pretrained plain vision transformer (Eqn.~\eqref{eqn:vit1}) to extract deep features from the support image and get $\mathbf{F}_s^*\in\mathbb{R}^{C\times H\times W}$, in which $H$ and $W$ are the feature height and width, and the subscript $s$ indicates the ``support''. Afterward, we use the support mask to crop out the foreground regions and background regions from $\mathbf{F}_s^*$. %Generating the foreground prompt is relatively simple: we average the cropped foreground features to get a foreground mean feature. This foreground mean feature is summed with several additional learnable tokens   Extracting the foreground. 
While the foreground regions are cropped out as a whole, the background regions are cropped and partitioned into multiple ($S$) local regions, because the background is likely not to be homogeneous. %To this end, we employ a Voronoi-based method \cite{aurenhammerVoronoi1991} to partition the background regions in three steps:
To this end, we employ a Voronoi-based method \cite{aurenhammerVoronoi1991} to partition the background into $S$ regions (See supplementary Section B.1 for details). %See supplementary for more details of selecting dispersed point and the Voronoi partition.

%\textbf{Step 1:} We collect all the background positions into a position set. 

%\textbf{Step 2:} We select $S$ dispersed points from the position set. These points should be as far as possible from each other. The selection procedure for such a dispersion effect is detailed in the \syf{supplementary}. 

%\textbf{Step 3:} Given these $S$ dispersed points as the seeds, we assign the neighboring pixels of each seed into a same part according to the Voronoi diagram (detailed in the \syf{supplementary}) and correspondingly derive $S$ local parts. 

According to the background partition results, we generate $S$ local masks for the background, \emph{i.e.}, $\{B_n\}_{n=1}^S$ with $B_n\in\{0,1\}^{H\times W}$. A partition example of $S=3$ is illustrated in Fig.~\ref{fig:methods}(a). Consequently, we calculate the mean feature of the foreground $\mathbf{u}_f^* \in \mathbb{R}^C$ and the local mean feature of background $\mathbf{u}_n^*\in\mathbb{R}^C$ by masked average pooling:
\begin{equation}\label{eqn:proxy}
    \mathbf{u}_f^* = \frac{1}{\vert \Tilde{Y}_s\vert}\sum_{i=1}^{HW}\mathbf{F}_{s,i}^*\Tilde{Y}_{s,i}, \quad \mathbf{u}_n^* = \frac{1}{\vert B_n\vert}\sum_{i=1}^{HW}\mathbf{F}_{s,i}^*B_{n,i}, \qquad n=1,\dots,S,
\end{equation}
where $\Tilde{Y}_s\in\{0,1\}^{H\times W}$ is the down-sampled foreground mask of the support image.  

% Given the downsampled support foreground mask $\Tilde{Y}_s\in\{0,1\}^{H\times W}$ and background masks $B_n,\;n=1,\dots,S$, the mean features $\mathbf{u}_f^*,\mathbf{u}_n^*\in\mathbb{R}^C$ can be calculated by mask average pooling:
% \begin{equation}\label{eqn:proxy}
%     \mathbf{u}_f^* = \frac{1}{\vert \Tilde{Y}_s\vert}\sum_{i=1}^{HW}\mathbf{F}_{s,i}^*\Tilde{Y}_{s,i}, \quad \mathbf{u}_n^* = \frac{1}{\vert B_n\vert}\sum_{i=1}^{HW}\mathbf{F}_{s,i}^*B_{n,i}, \qquad n=1,\dots,S.
% \end{equation}

Based on the mean features, we further expand each $C$-dimensional mean feature vector ($\mathbf{u}_f^* \in \mathbb{R}^C$ and $\mathbf{u}_n^*\in\mathbb{R}^C$) into a corresponding $G \times C$ token and then add it with extra learnable tokens by:
\begin{equation}
    \mathbf{p}_f = \mathcal{E}(\mathbf{u}_f^*) + \mathbf{z}_f, \quad
    \mathbf{p}_n = \mathcal{E}(\mathbf{u}_n^*) + \mathbf{z}_n, \qquad n=1,\dots,S, 
\end{equation}
where $\mathcal{E}$ is the expansion from $C$-dimensional vector to $G\times C$-dimensional token, and $\mathbf{z}_f,\mathbf{z}_n \in \mathcal{R}^{G\times C}$ are the learnable tokens. Adding these learnable tokens makes prompts gain extra diversity and become more discriminative  \cite{yangAssociating2021}. See supplementary Section B.2 for more details on this prompt augmentation technique. 
Consequently, after the expansion and augmentation, we get a foreground prompt token $\mathbf{p}_f \in \mathcal{R}^{G\times C}$ and $S$ local background prompt tokens $\mathbf{p}_n \in \mathcal{R}^{G\times C} (n=1,2, \cdots, S)$.

\subsubsection{Feature and Proxy Extraction}\label{sec:feature_proxy}

\paragraph{Feature Extraction.} The process for extracting features is illustrated in Fig.~\ref{fig:methods}(b). Without loss of generality, we take the 1-shot setting as the example for clarity. (See supplementary Section C for $K$-shot settings) Within each episode, a query image $I_q$ and a support image $I_s$ are split into $N$ patches and then flattened as $\{\mathbf{a}_{q,p}\}_{p=1}^N$ and $\{\mathbf{a}_{s,p}\}_{p=1}^N$, respectively.
An embedding layer projects these patches into query and support patch tokens, \emph{i.e.}, $\mathbf{X}_q^0\in\mathbb{R}^{N\times C}$ and $\mathbf{X}_s^0\in\mathbb{R}^{N\times C}$. 
We recall that in Section~\ref{sec:prompt}, we already get multiple prompt tokens, \emph{i.e.}, $\mathbf{P}^0=[\mathbf{p}_f, \mathbf{p}_0, \mathbf{p}_1, \cdots, \mathbf{p}_S]$. The patch tokens from a single image and the prompt tokens are concatenated as the query input and support input, \emph{i.e.}, $[\mathbf{X}_q^0, \mathbf{P}^0]$ and $[\mathbf{X}_s^0, \mathbf{P}^0]$. 
The query and support inputs are processed by the transformer blocks, which are formulated by:
\begin{align}\label{eqn:fptrans_1}
[\mathbf{X}_q^l,\mathbf{P}_q^l] &= B_l([\mathbf{X}_q^{l-1},\mathbf{P}^{l-1}]), %\qquad l=1,2,\dots,L, 
\\  \label{eqn:fptrans_2} 
[\mathbf{X}_s^l,\mathbf{P}_s^l] &= B_l([\mathbf{X}_s^{l-1},\mathbf{P}^{l-1}]),
%\qquad l=1,2,\dots,L
\\ \label{eqn:fptrans_3} 
\mathbf{P}^l &= (\mathbf{P}_q^l + \mathbf{P}_s^l)/2, 
\end{align}
where $l=1,2, \cdots, L$ enumerates all the transformer blocks. 

There is a novel and unique \textbf{prompt synchronization} (Eqn.~\eqref{eqn:fptrans_3}) in FPTrans. Specifically, Eqn.~\eqref{eqn:fptrans_3} averages the query and support prompt tokens after every transformer block. It makes the synced prompt token $\mathbf{P}_l$ absorb information from both the query and support patch tokens. Therefore, in the subsequential $(l+1)$-th block, the synced prompt token $\mathbf{P}_l$ passes the support information to the query tokens $[\mathbf{X}_q^{l+1}, \mathbf{P}_q^{l+1}]$ and vice versa. In a word, this simple prompt synchronization facilitates efficient interaction between query and support features.

\paragraph{Proxy Extraction.} As the results of Eqn.~\eqref{eqn:fptrans_1} to Eqn.~\eqref{eqn:fptrans_3}, the feature extractor outputs support features $\mathbf{X}_s^L$, query features $\mathbf{X}_q^L$, as well as the deep states of prompts $\mathbf{P}^L$. FPTrans uses the support features $\mathbf{X}_s^L$ and the deep states of prompts $\mathbf{P}^L$ to extract two types of proxies, \emph{i.e.}, the feature-based proxies and the prompt-based proxies, respectively.

$\bullet$ \emph{Feature-based proxies}. FPTrans uses the downsampled foreground mask $\Tilde{Y}_s$ and the partitioned background masks $B_n$ (which are already provided in Section~\ref{sec:prompt}) to extract the feature-based proxies by: $\mathbf{u}_f=\frac1{\vert \Tilde{Y}_s\vert}\sum_i\mathbf{F}_{s,i}\Tilde{Y}_{s,i},\;\mathbf{u}_n=\frac1{\vert B_n\vert}\sum_i\mathbf{F}_{s,i}B_{n,i}\;n=1,\dots,S$. 

$\bullet$ \emph{Prompt-based proxies}. Moreover, FPTrans uses the deep states of prompts $\mathbf{P}^L$ to get the prompt-based proxies by: $\mathbf{v}_f=\frac1G\sum_{j=1}^G\mathbf{p}_{f,j}^L$ and $\mathbf{v}_n=\frac1G\sum_{j=1}^G\mathbf{p}_{n,j}^L$.

These two types of proxies are both used for training FPTrans. In contrast, during testing, FPTrans only uses feature-based proxies. 

% The support and query features are reshaped back into 2D feature maps, denoted as $\mathbf{F}_s, \mathbf{F}_q \in \mathbb{R}^{C\times H\times W}$. To calculate proxies, we use the same partition strategy as Sec.~\ref{sec:prompt} to generate multiple local background regions. With the downsampled foreground mask $\Tilde{Y}_s$ and background partitions $B_n$, foreground and background proxies can be calculated $\mathbf{u}_f=\frac1{\vert \Tilde{Y}_s\vert}\sum_i\mathbf{F}_{s,i}\Tilde{Y}_{s,i},\;\mathbf{u}_n=\frac1{\vert B_n\vert}\sum_i\mathbf{F}_{s,i}B_{n,i}\;n=1,\dots,S$. When it comes to $K$-shot setting, we average the $K$ foreground proxies into a single one, and keep all the $KS$ background proxies. We also use the deep states of prompts $\mathbf{P}^L$ as extra proxies, which are calculated by averaging each prompt group of size $G$, resulting in prompt proxies $\mathbf{v}_f=\frac1G\sum_{j=1}^G\mathbf{p}_{f,j}^L$ and $\mathbf{v}_n=\frac1G\sum_{j=1}^G\mathbf{p}_{n,j}^L$. 

\subsubsection{Training and Inference}\label{sec:train_infer}
FPTrans uses two classification losses (one for the feature-based proxies and the other for the prompt-based proxies) and a pairwise loss for training.

\textbf{Classification losses.} Since segmentation can be viewed as a pixel-wise classification problem, FPTrans directly compares the similarity of each query feature vector ($\mathbf{F}_{q,i}$) and the proxies for linear classification. %We note that there are two sets of proxies, \emph{i.e.}, the feature-based and the prompt-based proxies. 
We elaborate on the classification loss with the feature-based proxy (and the process with the prompt-based proxies is similar). 
The predicted probability of $\mathbf{F}_{q,i}$ belonging to the foreground is formulated as:
\begin{equation}\label{eq:prob}
    \mathcal{P}(\mathbf{F}_{q,i}) = \frac{\exp({\texttt{sim}(\mathbf{F}_{q,i}, \mathbf{u}_f})/\tau)}
    {
    \exp{(\texttt{sim}(\mathbf{F}_{q,i}, \mathbf{u}_f)/\tau)}+
    \max_n(\exp{(\texttt{sim}(\mathbf{F}_{q,i}, \mathbf{u}_n/\tau)}))
    }, \quad n=1,2, \cdots, S
\end{equation}
where $\texttt{sim}(\cdot,\cdot)$ is the cosine similarity between two vectors, $\tau$ is a temperate coefficient. The $\max$ operation is critical because it facilitates comparing a feature to its closest background proxy. 

Given the predicted probability and the ground-truth label, FPTrans uses the standard cross-entropy loss $\mathcal{L}_{ce}$ for supervising the pixel-wise classification. In parallel to $\mathcal{L}_{ce}$, FPTrans uses the prompt-based proxies in a similar procedure and derives another classification loss $\mathcal{L}_{ce}'$. 

\textbf{Pairwise loss.} In addition to the classification losses, FPTrans further employs a pairwise loss to pull close all the foreground features from the query and support samples, as well as to push the foreground and background features far away from each other. The pairwise loss is formulated as:
\begin{equation}\label{eq:pair}
    \mathcal{L}_{pair} = \frac1{Z}\sum_{(Y_{q,i}+Y_{s,j})\geq 1}\text{BCE}(\sigma(\texttt{sim}(\mathbf{F}_{q,i}, \mathbf{F}_{s,j})/\tau), \mathbf{1}[Y_{q,i}=Y_{s,j}]),
\end{equation}
where $Z=\vert (Y_{q,i}+Y_{s,j})\geq 1\vert$ is the normalization factor and $\text{BCE}$ is the binary cross-entropy loss. $\sigma$ is a Sigmoid layer and $\mathbf{1}[\cdot]$ is the indicator function. %The intuition is that: if a query feature $\mathcal{F}_{q,i}$ and a support feature $\mathcal{F}_{s,j}$ both belong to the foreground, their cosine similarity should be enlarged to approach $1$. In contrary, 
If two features both belong to the background ($ (Y_{q,i}+Y_{s,j})=0)$, this pairwise loss function will NOT pull them close. Ablation studies show that pulling two foreground features close substantially improves FSS (Table~\ref{tab:ablation_components}) while pulling two background features actually compromises the FSS accuracy, as evidenced in the supplementary Section E.2.

\textbf{Overall, } FPTrans sums all the three losses for training:
\begin{equation}\label{eq:overall}
    \mathcal{L} = \mathcal{L}_{ce}+\mathcal{L}_{ce}' + \lambda\mathcal{L}_{pair},
\end{equation}
where $\lambda$ is a hyperparameter. 
For inference, FPTrans simply uses the query feature and the feature-based proxies for the final prediction (Eqn.~\eqref{eq:prob})

\section{Experiments}

\begin{table}[!t]
  \caption{Comparison with state-of-the-art methods on PASCAL-5$^i$. We report 1-shot and 5-shot results using the mean IoU (\%).}
  \label{tab:pascal}
  \centering
  \small
  \setlength\tabcolsep{5.4pt}%
  \begin{tabular}{clcccccccccc}
    \toprule
    \multirow{2}{*}{Backbone} & \multirow{2}{*}{Method} &  \multicolumn{5}{c}{1-shot} & \multicolumn{5}{c}{5-shot} \\
    \cmidrule{3-12}
      &  & S0 & S1 & S2 & S3 & Mean & S0 & S1 & S2 & S3 & Mean \\
    \midrule
    \multirow{5}{*}{Res-50} & RPMM~\cite{yangPrototype2020} & 55.2 & 66.9 & 52.6 & 50.7 & 56.3 & 56.3 & 67.3 & 54.5 & 51.0 & 57.3 \\
    % PPNet~\cite{liuPartAware2020} & & 47.8 & 58.8 & 53.8 & 45.6 & 51.5 & 58.4 & 67.8 & 64.9 & 56.7 & 62.0 \\
    & PFENet~\cite{tianPrior2020} & 61.7 & 69.5 & 55.4 & 56.3 & 60.8 & 63.1 & 70.7 & 55.8 & 57.9 & 61.9 \\
    & CyCTR~\cite{zhang2021few} & 67.8 & 72.8 & 58.0 & 58.0 & 64.2 & 71.1 & 73.2 & 60.5 & 57.5 & 65.6 \\
    & HSNet~\cite{minHypercorrelation2021} & 64.3 & 70.7 & 60.3 & 60.5 & 64.0 & 70.3 & 73.2 & 67.4 & 67.1 & 69.5 \\
    & BAM~\cite{langLearning2022} & 69.0 & 73.6 & 67.6 & 61.1 & 67.8 & 70.6 & 75.1 & 70.8 & 67.2 & 70.9 \\
    \midrule
    % FWB~\cite{Nguyen_2019_ICCV} & \multirow{6}{*}{Res-101} & 51.3 & 64.5 & 56.7 & 52.2 & 56.2 & 54.9 & 67.4 & 62.2 & 55.3 & 59.9 \\
    \multirow{5}{*}{Res-101} & DAN~\cite{wangFewShot2020a} & 54.7 & 68.6 & 57.8 & 51.6 & 58.2 & 57.9 & 69.0 & 60.1 & 54.9 & 60.5 \\
    & RePRI~\cite{boudiafFewShot2021} & 59.6 & 68.6 & 62.2 & 47.2 & 59.4 & 66.2 & 71.4 & 67.0 & 57.7 & 65.6 \\
    & PFENet~\cite{tianPrior2020} & 60.5 & 69.4 & 54.4 & 55.9 & 60.1 & 62.8 & 70.4 & 54.9 & 57.6 & 61.4 \\
    & CyCTR~\cite{zhang2021few} & 69.3 & 72.7 & 56.5 & 58.6 & 64.3 & 73.5 & 74.0 & 58.6 & 60.2 & 66.6 \\
    & HSNet~\cite{minHypercorrelation2021} & 67.3 & 72.3 & 62.0 & 63.1 & 66.2 & 71.8 & 74.4 & 67.0 & 68.3 & 70.4 \\
    \midrule
    \multirow{2}{*}{ViT-B/16} & Baseline & 62.9 & 69.1 & 62.2 & 53.0 & 61.8 & 70.5 & 76.0 & 74.2 & 65.5 & 71.5 \\
    & \cellcolor{grey1} \textbf{FPTrans} & \cellcolor{grey1} \bd67.1 & \cellcolor{grey1} \bd69.8 & \cellcolor{grey1} \bd65.6 & \cellcolor{grey1} \bd56.4 & \cellcolor{grey1} \bd64.7 & \cellcolor{grey1} \bd73.5 & \cellcolor{grey1} \bd75.7 & \cellcolor{grey1} \bd77.4 & \cellcolor{grey1} \bd68.3 & \cellcolor{grey1} \bd73.7 \\
    \midrule
    \multirow{2}{*}{DeiT-B/16} & Baseline & 68.2 & 69.4 & 61.7 & 60.5 & 64.9 & 75.3 & 78.1 & 76.1 & 73.7 & 75.8 \\
    & \cellcolor{grey1} \textbf{FPTrans} & \cellcolor{grey1} \bd72.3 & \cellcolor{grey1} \bd70.6 & \cellcolor{grey1} \bd68.3 & \cellcolor{grey1} \bd64.1 & \cellcolor{grey1} \bd68.8 & \cellcolor{grey1} \bd76.7 & \cellcolor{grey1} \bd79.0 & \cellcolor{grey1} \bd81.0 & \cellcolor{grey1} \bd75.1 & \cellcolor{grey1} \bd78.0 \\
    \bottomrule
  \end{tabular}
\end{table}

\subsection{Implementation Details}\label{sec:imple}

\paragraph{Datasets and Metrics.} We use two popular FSS benchmarks PASCAL-5$^i$~\cite{shaban2017one} and COCO-20$^i$~\cite{tianPrior2020} for evaluation. PASCAL-5$^i$ combines PASCAL VOC 2012~\cite{pascal-voc-2012} and SBD~\cite{hariharanSimultaneous2014}, and includes 20 classes. Following prior works~\cite{tianPrior2020,langLearning2022}, we split the dataset into four splits with each split 15 classes for training and 5 classes for testing. COCO-20$^i$ is constructed with COCO 2014~\cite{linMicrosoft2014} and includes 80 classes. It is divided into 4 splits with each split 60 classes for training and 20 classes for testing. To compare with previous methods, we report mean IoU~(mIoU) averaged on test classes~\cite{tianPrior2020,zhang2021few,langLearning2022}.

\paragraph{Training Details.} All the images are resized and cropped to $480\times 480$ and augmented following~\cite{tianPrior2020}. We evaluate the proposed method on two vision transformer backbones, ViT-B/16~\cite{dosovitskiyImage2021} and DeiT-B/16~\cite{touvronTraining2021}. These two backbones are both pretrained on Imagenet-1k~\cite{dengImageNet2009}. The cross-entropy losses are optimized with boundary-enhanced weight maps introduced by~\cite{zhang2021PriorEnhanced}. We use the SGD optimizer with a momentum of 0.9, a weight decay of 5e-5, and a constant learning rate of 1e-3. Using 4 A100 GPUs, we train 60 epochs with ViT and 30 epochs with DeiT backbone, using a batch size 4 for PASCAL-5$^i$ and 16 for COCO-20$^i$ (batch size 8 in 5-shot due to the memory limitation). When we generate the local background prompts (and the feature-based proxies), the background of each support image is partitioned into $5$ local parts, \emph{i.e.}, $S=5$. Each prompt consists of $12$ tokens, \emph{i.e.}, $G=12$.  The weight factor $\lambda$ for balancing the classification loss and pairwise loss (Eqn.~\eqref{eq:overall}) is set as 2e-2 for PASCAL-5$^i$ and 1e-4 for COCO-20$^i$. Our baseline is implemented as the plain vision transformer.

\begin{table}[!t]
  \caption{Comparison with state-of-the-art methods on COCO-20$^i$. We report 1-shot and 5-shot results using the mean IoU (\%).}
  \label{tab:coco}
  \centering
  \small
  \setlength\tabcolsep{5.4pt}%
  \begin{tabular}{clcccccccccc}
    \toprule
    \multirow{2}{*}{Backbone} & \multirow{2}{*}{Method} & \multicolumn{5}{c}{1-shot} & \multicolumn{5}{c}{5-shot} \\
    \cmidrule{3-12}
      &  & S0 & S1 & S2 & S3 & Mean & S0 & S1 & S2 & S3 & Mean \\
    \midrule
    \multirow{3}{*}{Res-50} & RePRI~\cite{boudiafFewShot2021} & 32.0 & 38.7 & 32.7 & 33.1 & 34.1 & 39.3 & 45.4 & 39.7 & 41.8 & 41.6 \\
    & HSNet~\cite{minHypercorrelation2021} & 36.3 & 43.1 & 38.7 & 38.7 & 39.2 & 43.3 & 51.3 & 48.2 & 45.0 & 46.9 \\
    & BAM~\cite{wangLearning2022} & 43.4 & 50.6 & 47.5 & 43.4 & 46.2 & 49.3 & 54.2 & 51.6 & 49.6 & 51.2 \\
    \midrule
    \multirow{3}{*}{Res-101} & DAN~\cite{wangFewShot2020a} & - & - & - & - & 24.4 & - & - & - & - & 29.6 \\
    & PFENet~\cite{tianPrior2020} & 34.3 & 33.0 & 32.3 & 30.1 & 32.4 & 38.5 & 38.6 & 38.2 & 34.3 & 37.4 \\
    & HSNet~\cite{minHypercorrelation2021} & 37.2 & 44.1 & 42.4 & 41.3 & 41.2 & 45.9 & 53.0 & 51.8 & 47.1 & 49.5 \\
    \midrule
    \multirow{2}{*}{ViT-B/16} & Baseline & 37.3 & 39.6 & 41.5 & 35.3 & 38.4 & 48.2 & 53.5 & 52.9 & 48.8 & 50.8 \\
    & \cellcolor{grey1} \textbf{FPTrans} & \cellcolor{grey1} \cellcolor{grey1} \bd39.7 & \cellcolor{grey1} \bd44.1 & \cellcolor{grey1} \bd44.4 & \cellcolor{grey1} \bd39.7 & \cellcolor{grey1} \bd42.0 & \cellcolor{grey1} \bd49.9 & \cellcolor{grey1} \bd56.5 & \cellcolor{grey1} \bd55.4 & \cellcolor{grey1} \bd53.2 & \cellcolor{grey1} \bd53.8 \\
    \midrule
    \multirow{2}{*}{DeiT-B/16} & Baseline & 41.8 & 45.4 & 48.8 & 40.3 & 44.1 & 53.9 & 60.1 & 58.9 & 54.4 & 56.8  \\
    & \cellcolor{grey1} \textbf{FPTrans} & \cellcolor{grey1} \bd44.4 & \cellcolor{grey1} \bd48.9 & \cellcolor{grey1} \bd50.6 & \cellcolor{grey1} \bd44.0 & \cellcolor{grey1} \bd47.0 & \cellcolor{grey1} \bd54.2 & \cellcolor{grey1} \bd62.5 & \cellcolor{grey1} \bd61.3 & \cellcolor{grey1} \bd57.6 & \cellcolor{grey1} \bd58.9 \\
    \bottomrule
  \end{tabular}
\end{table}

\begin{table}[!t]
\begin{minipage}[t]{0.49\textwidth}
% \makeatletter\def\@captype{table}
    \caption{Evaluation (Mean IoU (\%)) under the domain shift from COCO-20$^i$ to PASCAL-5$^i$.}
    \label{tab:domain_shift}
    \centering
    \small
    \setlength\tabcolsep{4pt}
    \vspace{3pt}
    \begin{tabular}{cccc}
        \toprule
        \multirow{2}{*}{Method} & \multirow{2}{*}{Backbone} & \multicolumn{2}{c}{COCO$\rightarrow$PASCAL}  \\
        % \cmidrule{3-4}
         & & 1-shot & 5-shot \\
        \midrule
        PFENet~\cite{tianPrior2020} & \multirow{2}{*}{Res-50} & 61.1 & 63.4 \\
        RePRI~\cite{boudiafFewShot2021} &  & 63.2 & 67.7 \\
        % HSNet~\cite{minHypercorrelation2021} &  & 61.6 & 68.7 \\
        \midrule
        HSNet~\cite{minHypercorrelation2021} & Res-101 & 64.1 & 70.3 \\
        \midrule
        \multirow{2}{*}{FPTrans} & ViT-B/16 & 67.6 & 76.9 \\
         & DeiT-B/16 & \bd69.7 & \bd79.3 \\
        \bottomrule
    \end{tabular}
\end{minipage}
\hfill
\begin{minipage}[t]{0.49\textwidth}
% \makeatletter\def\@captype{table}
    \caption{Ablation studies. ``Pair Loss'', ``Prompts'' and ``Proxies'' control using (or not using) the pairwise loss, the prompts, and the multiple local background proxies, respectively. }
    \label{tab:ablation_components}
    \centering
    \small
    \setlength\tabcolsep{3pt}
    \begin{tabular}{ccccc}
        \toprule
        Pair Loss & Prompts    & Proxies & PASCAL & COCO  \\
        \midrule
                       &            &                  & 61.8         & 38.4         \\
        \checkmark     &            &                  & 62.9         & 38.8         \\
        \checkmark     & \checkmark &                  & 63.9         & 41.5         \\
        \checkmark     &            & \checkmark       & 64.0         & 40.3         \\
        \checkmark     & \checkmark & \checkmark       & \bd64.7      & \bd42.0         \\
        \bottomrule
    \end{tabular}
\end{minipage}
\end{table}

\subsection{Comparison with State-of-the-Art Methods}

\paragraph{Main results.} Table~\ref{tab:pascal} evaluates the proposed FPTrans  on PASCAL-5$^i$, from which we draw two observations as follows:

First, FPTrans consistently improves the baseline on two backbones. For example, when using the ViT-B/16 as its baseline, FPTrans surpasses the baseline by +2.9\% and +2.2\% mIoU on 1-shot and 5-shot settings, respectively. We note that compared with the baseline, FPTrans has three major differences, \emph{i.e.}, query-support interactions to better utilize the limited support samples, multiple local background proxies to promote novel-class generalization, and an additional pairwise loss function to pull close foreground features. Ablation studies in Table~\ref{tab:ablation_components} confirm these advantages as the main reasons that improve the baseline. 

Second, comparing FPTrans against state-of-the-art methods, we find that FPTrans achieves competitive FSS accuracy. Under the 1-shot setting, FPTrans on DeiT-B/16 surpasses the most competitive BAM~\cite{langLearning2022} by 1.0\% mIoU. Under the 5-shot setting, the superiority of FPTrans is even larger, \ie, +2.8\% based on ViT-B/16 and +7.1\% based on DeiT-B/16.

%comparing with the ViT-B/16 baseline, FPTrans achieves +2.9\% and +2.2\% mIoU gains on 1-shot and 5-shot settings. Second, with a DeiT-B/16 backbone, FPTrans improves the baseline by +3.9\% and +2.2\% mIoU gains on the two settings. Third, FPTrans with DeiT-B/16 achieves state-of-the-art results on both 1-shot and 5-shot settings, in which FPTrans significantly outperforms the competitive BAM~\cite{langLearning2022} by +2.8\% and +7.1\% with ViT-B/16 and DeiT-B/16, respectively, on the 5-shot setting.
Table~\ref{tab:coco} summarizes the results on COCO-20$^i$. The major observations are consistent as on PASCAL-5$^i$. FPTrans on ViT-B/16 and DeiT-B/16 both surpass the prior state of the art by a clear margin, setting a new state of the art. 
%We observe that FPTrans with the ViT-B/16 improves the baseline by more than +3.0\% on both 1-shot and 5-shot settings. Moreover, FPTrans with DeiT-B/16 outperforms the competitive BAM by +0.8\% and +7.7\% on the 1-shot and 5-shot setting, achieving state-of-the-art results. 

\paragraph{Domain shift scenario.} Few-shot learning has been studied under a domain shift scenario~\cite{hu2022switch}. Therefore, we also evaluate the proposed FPTrans on the domain shift scenario for semantic segmentation, where the base training data and the testing data have a significant domain gap. 
%FPTrans can handle the FSS problem under a domain shift scenario. 
We use COCO-20$^i$ for training and use PASCAL-5$^i$ for testing, following previous works ~\cite{boudiafFewShot2021,minHypercorrelation2021} for comparison. The training classes (in COCO-20$^i$) and the novel testing classes (in PASCAL-5$^i$) are not overlapped. We summarize the average results on 4 COCO-trained models in Table~\ref{tab:domain_shift}. FPTrans outperforms HSNet~\cite{minHypercorrelation2021} by +5.6\% and +9.0\% on the 1-shot and 5-shot settings, respectively. It confirms the effectiveness of FPTrans under the domain shift scenario.

\subsection{Ablation Study}\label{sec:ablation}
To better understand the proposed methods, we conduct ablation studies on FPTrans components. Experiments are conducted with a ViT-B/16 backbone on the 1-shot setting unless specified otherwise. 

\paragraph{Ablations on some major components.} We recall that FPTrans has multiple keypoints / important designs, \emph{i.e.}, an additional pairwise loss for training, a novel prompting strategy, and multiple local proxies. The corresponding ablations are shown in Table~\ref{tab:ablation_components}, from which we draw three observations. First, the pairwise loss improves the baseline by +1.1\% gains on PASCAL-5$^i$ and +0.4\% on COCO-20$^i$. Second, adding prompts further brings +1.0\% and +2.7\% gains while adding multiple background proxies further brings +1.1\% and +1.5\% gains on PASCAL-5$^i$ and COCO-20$^i$, respectively. Third, compared with the baseline, the full FPTrans equipped with all the three components achieves overall improvements of +2.9\% and 3.6\% on PASCAL-5$^i$ and COCO-20$^i$, respectively. %It testifies the effectiveness of the FPTrans.

\begin{table}[!t]
\begin{minipage}[t]{0.45\textwidth}
    \caption{Performance (Mean IoU (\%)) of previous methods with transformer backbones. Experiments are conducted on PASCAL-5$^i$.}
    \label{tab:backbone}
    \centering
    \small
    \setlength\tabcolsep{3pt}
    \begin{tabular}{ccccc}
        \toprule
        Method & Res-50 & Res-101 & ViT & DeiT \\
        \midrule
        PFENet~\cite{tianPrior2020} & 60.8 & 60.1 & 58.7 & 57.7 \\
        CyCTR~\cite{zhang2021few} & 64.2 & 64.3 & 60.1 & 61.0 \\
        \midrule
        Baseline                  & - & - & 61.8 & 64.9 \\
        FPTrans                   & - & - & \bd64.7 & \bd68.8 \\
        \bottomrule
    \end{tabular}
\end{minipage}
\hfill
\begin{minipage}[t]{0.53\textwidth}
    \caption{Results (Mean IoU (\%)) of the different number of background proxies $S$ with the proposed methods (Ours) or mixture model~(Mix.)~\cite{yangPrototype2020}.}
    \label{tab:bg_proxy}
    \centering
    \small
    \setlength\tabcolsep{4pt}
    \begin{tabular}{ccccccc}
        \toprule
        \multicolumn{2}{c}{$ S $} & 1 & 3 & 5 & 7 & 9 \\
        \midrule
         \multirow{2}{*}{PASCAL-5$^i$} & Ours    & 63.9 & 64.4 & \bd64.7 & 64.4 & 64.0 \\
                                 & Mix. & -    & 64.2 & 64.1 & -    & -    \\ 
        \midrule
         \multirow{2}{*}{COCO-20$^i$}   & Ours    & 41.5 & 41.9 & \bd42.0 & 41.7 & 41.2 \\
                                 & Mix. & -    & 40.9 & 40.6 & -    & -    \\
        \bottomrule
    \end{tabular}
\end{minipage}
\end{table}

\paragraph{Transformer backbone for decoder-based method.} Based on two competitive decoder-based methods PFENet~\cite{tianPrior2020} and CyCTR~\cite{zhang2021few}, we replace their CNN backbones with the ViT-B/16 and DeiT-B/16 backbones, as shown in Table~\ref{tab:backbone}. We observe that these two methods undergo considerable accuracy decreases after the backbone replacement. It suggests that replacing the CNN backbone with a transformer does not necessarily improve FSS, although the transformer backbone fits our plain FSS framework. 

\begin{wrapfigure}{r}{0.55\linewidth}
    \centering
    \vspace{-15pt}
    \includegraphics[width=\linewidth]{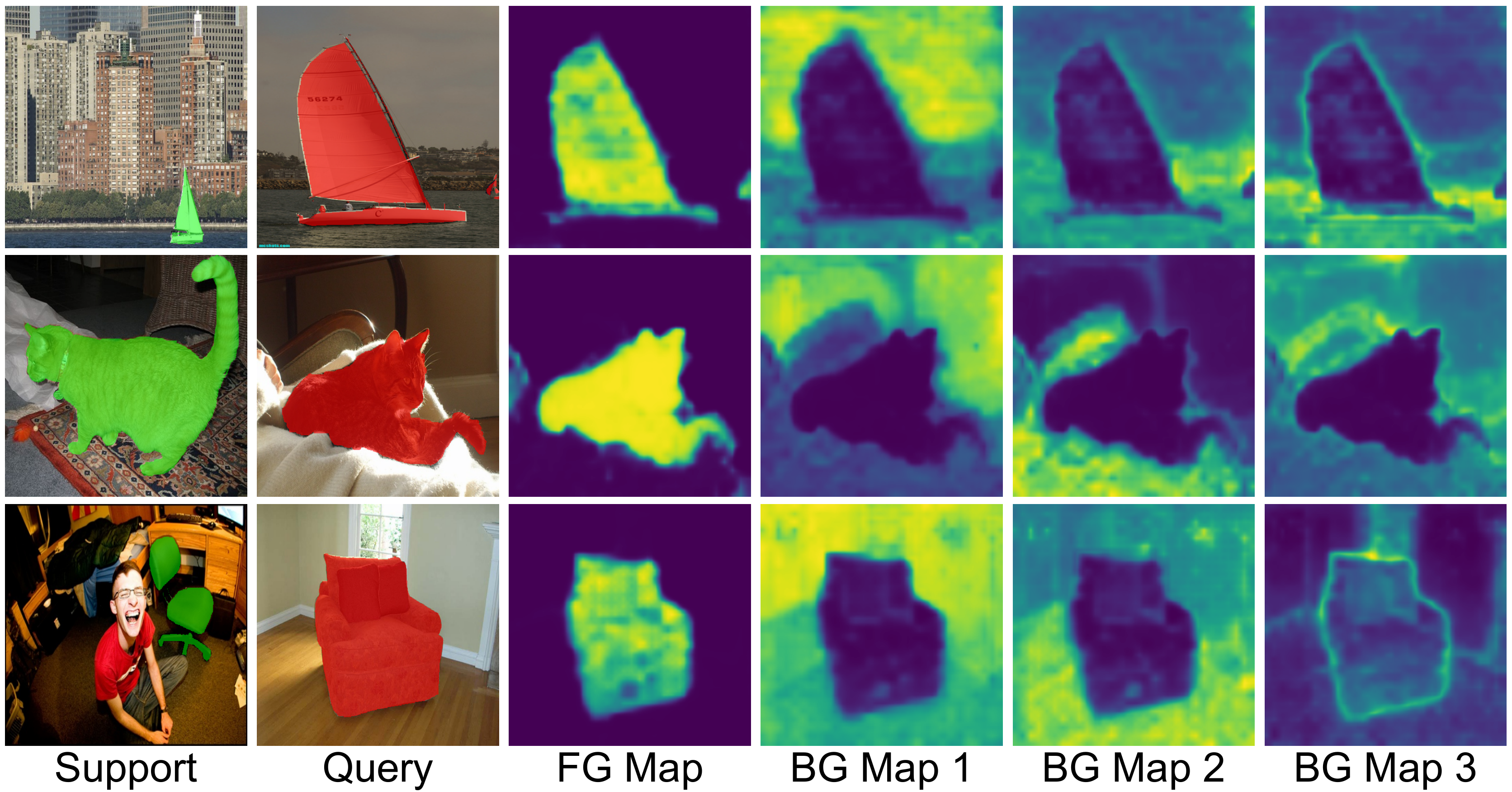}
    \vspace{-10pt}
    \caption{Visualization of predicted score maps. Each background proxy focus on the partial background.}
    \label{fig:vis}
    \vspace{-15pt}
\end{wrapfigure}

\paragraph{Investigation of the local background proxies. } We investigate the local background proxies by varying their numbers $S$ in Table~\ref{tab:bg_proxy}. It is observed that using 5 background proxies achieves the highest accuracy on both PASCAL-5$^i$ and COCO-20$^i$ for our method. 
Moreover, we compare our method with another multi-proxy method (``Mix.'') proposed in RPMM~\cite{yangPrototype2020}. RPMM online trains mixture models to generate multiple background proxies. In contrast, our approach for achieving multiple background is relatively simple (by partitioning the background) and superior (\emph{e.g.}, +0.6\% gains on PASCAL-5$^i$ and +1.4\% gains on COCO-20$^i$ when $S=5$. We infer that our partition-based background proxies are more representative because it considers a realistic factor, \emph{i.e.}, the far-away regions of the background are likely to be inhomogeneous to each other. 
Fig.~\ref{fig:vis} visualizes the activation maps of different proxies (one foreground proxy and three local background proxies). It confirms that the local background proxies well accommodate the inhomogeneous background. 
%The visualization in Fig.~\ref{fig:vis} also confirms this conclusion, \ie, each background proxy mainly focus on part of the background.

\section{Conclusion}\label{sec:conclusion}

This paper revives the plain framework ``feature extractor $+$ linear classification head'' and correspondingly proposes a novel Feature-Proxy Transformer~(FPTrans) for few-shot segmentation. During feature extraction, FPTrans makes the query interact with support features in all the transformer blocks, therefore well utilizing the limited support samples. During proxy extraction, FPTrans encodes the complex background into multiple local background proxies, therefore improving the generalization towards novel classes. Given the discriminative features and the representative proxies, FPTrans directly uses a linear classification head to compare their cosine similarity and achieves state-of-the-art performance. With its simplicity and competitive accuracy, we hope FPTrans can serve as a strong baseline for few-shot segmentation.  

% FPTrans utilizes the limited support samples by making the query interact with support features across the feature extractor with a novel prompting strategy. Moreover, FPTrans uses multiple local background proxies to encode the inhomogeneous background regions for more representative proxies. Without using intricate decoders, FPTrans directly compares the similarity of the adaptive features and proxies for segmentation. Despite its simplicity, FPTrans achieves competitive accuracy comparing with the state-of-the-art methods, and we hope it can serve as a strong baseline for future FSS works.

\vspace{-2pt}
\paragraph{Limitations and future works.} %Although FPTrans achieves promising results with a simple framework, it r
Currently, FPTrans relies on a pretrained model for generating prompts, which consumes extra computation resources. However, we find that FPTrans is robust to the model for extracting the prompts (see supplementary Section F for details) to some extent. Therefore, we will seek more lightweight prompt-generating models in future works. 

\section*{Acknowledgements}

The work of the authors was supported by the National Natural Science Foundation of China (No. 62132017), and the Fundamental Research Funds for the Central Universities (No. 226-2022-00087).

{
\small
\bibliographystyle{plain}
\bibliography{neurips_2022}
}

%%%%%%%%%%%%%%%%%%%%%%%%%%%%%%%%%%%%%%%%%%%%%%%%%%%%%%%%%%%%
\section*{Checklist}

%%% BEGIN INSTRUCTIONS %%%
% The checklist follows the references.  Please
% read the checklist guidelines carefully for information on how to answer these
% questions.  For each question, change the default \answerTODO{} to \answerYes{},
% \answerNo{}, or \answerNA{}.  You are strongly encouraged to include a {\bf
% justification to your answer}, either by referencing the appropriate section of
% your paper or providing a brief inline description.  For example:
% \begin{itemize}
%   \item Did you include the license to the code and datasets? \answerYes{See Section~\ref{gen_inst}.}
%   \item Did you include the license to the code and datasets? \answerNo{The code and the data are proprietary.}
%   \item Did you include the license to the code and datasets? \answerNA{}
% \end{itemize}
% Please do not modify the questions and only use the provided macros for your
% answers.  Note that the Checklist section does not count towards the page
% limit.  In your paper, please delete this instructions block and only keep the
% Checklist section heading above along with the questions/answers below.
%%% END INSTRUCTIONS %%%

\begin{enumerate}

\item For all authors...
\begin{enumerate}
  \item Do the main claims made in the abstract and introduction accurately reflect the paper's contributions and scope?
    \answerYes{}
  \item Did you describe the limitations of your work?
    \answerYes{See Section~\ref{sec:conclusion} and supplementary.}
  \item Did you discuss any potential negative societal impacts of your work?
    \answerYes{See Section~\ref{sec:conclusion}.}
  \item Have you read the ethics review guidelines and ensured that your paper conforms to them?
    \answerYes{}
\end{enumerate}

\item If you are including theoretical results...
\begin{enumerate}
  \item Did you state the full set of assumptions of all theoretical results?
    \answerNA{}
        \item Did you include complete proofs of all theoretical results?
    \answerNA{}
\end{enumerate}

\item If you ran experiments...
\begin{enumerate}
  \item Did you include the code, data, and instructions needed to reproduce the main experimental results (either in the supplemental material or as a URL)?
    \answerYes{See the supplementary.}
  \item Did you specify all the training details (e.g., data splits, hyperparameters, how they were chosen)?
    \answerYes{See Section~\ref{sec:imple} and Section~\ref{sec:ablation}.}
  \item Did you report error bars (e.g., with respect to the random seed after running experiments multiple times)?
    \answerYes{See the supplementary.}
  \item Did you include the total amount of compute and the type of resources used (e.g., type of GPUs, internal cluster, or cloud provider)?
    \answerYes{See Section~\ref{sec:imple}.}
\end{enumerate}

\item If you are using existing assets (e.g., code, data, models) or curating/releasing new assets...
\begin{enumerate}
  \item If your work uses existing assets, did you cite the creators?
    \answerYes{See Section~\ref{sec:imple}.}
  \item Did you mention the license of the assets?
    \answerYes{See the supplementary.}
  \item Did you include any new assets either in the supplemental material or as a URL?
    \answerNA{}
  \item Did you discuss whether and how consent was obtained from people whose data you're using/curating?
    \answerNA{}
  \item Did you discuss whether the data you are using/curating contains personally identifiable information or offensive content?
    \answerYes{See the supplementary.}
\end{enumerate}

\item If you used crowdsourcing or conducted research with human subjects...
\begin{enumerate}
  \item Did you include the full text of instructions given to participants and screenshots, if applicable?
    \answerNA{}
  \item Did you describe any potential participant risks, with links to Institutional Review Board (IRB) approvals, if applicable?
    \answerNA{}
  \item Did you include the estimated hourly wage paid to participants and the total amount spent on participant compensation?
    \answerNA{}
\end{enumerate}

\end{enumerate}

\newpage
\appendix

\section{A Detailed Revisit to Vision Transformer}
Vision transformer~(ViT)~\cite{dosovitskiyImage2021} derived from Transformer~\cite{vaswaniAttention2017} is designed for computer vision tasks. It consists of an embedding module, a sequence of stacked transformer blocks, and an MLP head. 
Specifically, given an RGB image $I$, ViT first crops and reshapes it into a series of image patches $\{\mathbf{a}_p\in\mathbb{R}^{3\times P\times P}\}_{p=1}^N$ ($P$ is the patch size), and then projects them into $C$-dimensional embeddings:
\begin{equation}
    \mathbf{x}_p = \mathtt{Embed}(\mathbf{a}_p), \qquad p=1,2,\dots, N.
\end{equation}
We denote the collection of embeddings as $\mathbf{X} = \{\mathbf{x}_p\}_{p=1}^N$.

In each transformer block $B_l$, the layer inputs are processed by a Multiheaded Self-Attention~(MSA) module and a multilayer perceptron ~(MLP) module with extra residual connections. More concretely, given a sequence of input tokens $\mathbf{X}\in\mathbb{R}^{N\times C}$, a self-attention layer corresponds to three matrices: queries $\mathbf{Q}=\mathbf{XW}^Q$, keys $\mathbf{K}=\mathbf{XW}^K$ and values $\mathbf{V}=\mathbf{XW}^V$, where $\mathbf{W}^Q, \mathbf{W}^K\in\mathbb{R}^{C\times d_k}$, and $\mathbf{W}^V\in\mathbb{R}^{C\times d_v}$ are projection weights. With $\mathbf{Q}, \mathbf{K}$ and $\mathbf{V}$, self-attention can be formulated as follows:
\begin{equation}
    \textrm{Attention}(\mathbf{Q}, \mathbf{K}, \mathbf{V}) = \textrm{Softmax}(\frac{\mathbf{QK}^{\top}}{\sqrt{d_k}})\mathbf{V}.
\end{equation}
MSA is constructed based on Attention by split the channels of $\mathbf{Q},\mathbf{K}$ and $\mathbf{V}$ into $h$ groups with each group a part of queries, keys, and values $\mathbf{Q}_i, \mathbf{K}_i\in\mathbb{R}^{N\times \frac{d_k}{h}}, \mathbf{V}_i\in\mathbb{R}^{N\times \frac{d_v}{h}}$. Therefore, MSA concatenates the attentions results of each group by:
\begin{equation}
    \textrm{MSA}(\mathbf{Q}, \mathbf{K}, \mathbf{V}) = [\textrm{head}_1,\textrm{head}_2,\cdots,\textrm{head}_h]\mathbf{W}^O,
\end{equation}
where $\textrm{head}_i=\textrm{Attention}(\mathbf{Q}_i, \mathbf{K}_i, \mathbf{V}_i)$, and $[\cdot,\cdot]$ is the concatenate operation. $\mathbf{W}^O\in\mathbb{R}^{d_v\times C}$ is the projection weights for layer outputs. 

With the MSA module and MLP module (a two-layer MLP), the ViT can be formulated as:
\begin{align} \label{eq:vit_1}
    \mathbf{X}_0 &= [\mathbf{x}_{cls}, \mathbf{X}_{img}] + \mathbf{E}_{pos}, \\ \label{eq:vit_2}
    \mathbf{X}_l' &= \textrm{MSA}(\textrm{LN}(\mathbf{X}_{l-1})) + \mathbf{X}_{l-1}, \qquad l=1,2,\dots, L, \\ \label{eq:vit_3}
    \mathbf{X}_l &= \textrm{MLP}(\textrm{LN}(\mathbf{X}_l')) + \mathbf{X}_l', \qquad l=1,2,\dots, L, \\
    \mathbf{y} &= \textrm{MLP\_Head}(\textrm{LN}(\mathbf{X}_L^0)), 
\end{align}
where $\mathbf{E}_{pos}\in\mathbb{R}^{(N+1)\times C}$ is the position embedding and LN is the layer normalization~\cite{BaevskiAdaptive2019}. MLP\_Head is a linear classifier and $\mathbf{X}_L^0$ is the deep state of the class token $\mathbf{x}_{cls}$. We note that Eqn.~\eqref{eq:vit_2} and Eqn.~\eqref{eq:vit_3} comprise the transformer block $B_l$.  

\section{Details for Prompt Generation}
As illustrated in Section \textcolor{red}{3.3.2} in the manuscript, we generate a set of prompts from each support image, including a foreground prompt and multiple local background prompts. Generating these local background prompts requires partitioning the background into several local parts. Moreover, we use an augmentation method and derive multiple prompt tokens from each generated prompt. The augmented prompt tokens are then concatenated with the patch tokens and are fed into the transformer blocks.
Here we provide some important details for the prompt generation procedure in the proposed FPTrans, \emph{i.e.}, the Voronoi-Based Background partition, and the prompt augmentation. 
\subsection{Voronoi-Based Background Partition}
FPTrans adopts a Voronoi-based method~\cite{aurenhammerVoronoi1991} to partition the background into multiple regions. The partition method consists of three steps:

\textbf{Step 1:} We collect all the background positions into a position set $\mathcal{B}=\{t=(i,j)\vert Y_{s,t}=0,\;i=1,2,\dots,W,\; j=1,2,\dots,H\}$, where $Y_s$ is the support mask, and $H$ and $W$ are the height and width, respectively. An empty point list $\mathcal{T}$ is initialized.

\textbf{Step 2:} We select $S$ dispersed points from the position set $\mathcal{B}$. Specifically, a seed $t_1$ is first randomly sampled from $\mathcal{B}$ and appended into $\mathcal{T}$. Then, we select from $\mathcal{B}$ the next seed point $t_2$ satisfying:
\begin{equation}\label{eq:select_t}
    t_2 = \underset{t\in\mathcal{B}}{\arg\max}\min_{t'\in\mathcal{T}}\Vert t - t'\Vert_2^2,
\end{equation}
which is the farthest point to all the points in $\mathcal{T}$. $t_2$ is appended into $\mathcal{T}$. Repeating Eqn.~\eqref{eq:select_t}, we can finally select $S$ points dispersed in the background.

\textbf{Step 3:} Given these $S$ dispersed points $\mathcal{T}=\{t_1,t_2,\dots,t_S\}$ as the seeds, we assign the neighboring pixels of each seed into the same part according to the Voronoi diagram and correspondingly derive $S$ local parts. Concretely, for each background pixel $t\in\mathcal{B}$, we assign a label $m_t$ as follows:
\begin{equation}
    m_t = \underset{n\in\{1,2,\dots,S\}}{\arg\min}\Vert t - t_n\Vert_2^2.
\end{equation}
Therefore, the background is divided into $S$ regions formulated as $\mathcal{B}_n=\{t\in\mathcal{B}\vert m_t=n\},\;n=1,2,\dots,S$. The point sets $\mathcal{B}_n$ are further transferred into binary masks as $B_n=\{0,1\}^{H\times W}$ for the subsequent masked average pooling as stated in the Section~3.3.2 of the main manuscript.

\subsection{Prompt Augmentation} 
With the downsampled support foreground mask $\Tilde{Y}_s$ and background partitions $B_n,\;n=1,2,\dots,S$, we calculate the foreground and multiple background mean features by masked average pooling:
\begin{equation}
    \mathbf{u}_f^*=\frac1{\vert\Tilde{Y}_s\vert}\sum_{i=1}^{H\times W}\mathbf{F}_{s,i}^*\Tilde{Y}_{s,i},\quad \mathbf{u}_n^*=\frac1{\vert B_n\vert}\sum_{i=1}^{H\times W}\mathbf{F}_{s,i}^*B_{s,i}, \quad n=1,2,\dots,S,
\end{equation}
where $\mathbf{F}_s^*$ is the support features extracted by a pretrained ViT. Inspired by the multiple-object tracking within a single framework~\cite{yangAssociating2021}, in which different objects are represented by various identifications~(\ie, learnable vectors) for simultaneously tracking, we add extra learnable tokens to the mean features for more discriminative prompts. Specifically, we first expand $C$-dimensional mean features ($\mathbf{u}_f$ and $\mathbf{u}_n$) into a corresponding token $\mathcal{E}(\mathbf{u}_f),\mathcal{E}(\mathbf{u}_n)\in\mathbb{R}^{G\times C}$ (by repeating each mean feature $G$ times). Then, a group of learnable tokens is added to obtain the final prompts:
\begin{equation}\label{eq:prompts}
    \mathbf{p}_f=\mathcal{E}(\mathbf{u}_f^*)+\mathbf{z}_f, \quad
    \mathbf{p}_n=\mathcal{E}(\mathbf{u}_n^*)+\mathbf{z}_n, \quad n=1,2,\dots,S.
\end{equation}
We note that in different episodes, $\mathbf{u}_f^*$ and $\mathbf{u}_n^*$ represent diverse foreground and background classes, which implies that $\mathbf{z}_f$ and $\mathbf{z}_n$ should not be bound to specific classes. To this end, we initialize a learnable token pool $\mathcal{W}=\{\mathbf{z}\vert \mathbf{z}\in\mathbb{R}^{G\times C}\}$ with size $\vert\mathcal{W}\vert=D$. In each episode, $S+1$ learnable tokens are randomly sampled $\{\mathbf{z}_f,\mathbf{z}_1,\mathbf{z}_2,\dots,\mathbf{z}_S\}\subset\mathcal{W}$ and used for constructing prompts using Eqn.~\eqref{eq:prompts}. In this way, these tokens are optimized (by gradient back-propagation from foreground and background prompts) to be diverse from each other, which in turn enhances the discrimination of prompts.

\section{Details for K-Shot Setting}
FPTrans can be naturally extended to the $K$-shot setting when $K>1$. 

Specifically, for the \textbf{prompt generation}~(the process of \textbf{feature-based proxy extraction} is similar), we calculate a foreground mean feature and $S$ background mean features for each support sample by:
\begin{align}
    \mathbf{u}_f^{(k)} &= \frac{1}{\vert \Tilde{Y}_s^{(k)}\vert}\sum_{i=1}^{HW}\mathbf{F}_{s,i}^{(k)}\Tilde{Y}_{s,i}^{(k)}, \quad k=1,\dots,K \\
    \mathbf{u}_n^{(k)} &= \frac{1}{\vert B_n^{(k)}\vert}\sum_{i=1}^{HW}\mathbf{F}_{s,i}^{(k)}B_{n,i}^{(k)}, \quad k=1,\dots,K,\;\; n=1,\dots,S,
\end{align}
where $\mathbf{F}_s^{(k)}$ is the support feature, $\Tilde{Y}_s^{(k)}\in\{0,1\}^{H\times W}$ is the down-sampled foreground mask, and $\{B_n^{(k)}\}_{n=1}^S$ are background partitions. The final foreground mean feature (or proxy) is calculated by taking the average on $\{\mathbf{u}_f^{(k)}\}_{k=1}^K$ following prior methods~\cite{tianPrior2020,zhang2021few}:
\begin{equation}
    \mathbf{u}_f = \frac1K\sum_{k=1}^K\mathbf{u}_f^{(k)}.
\end{equation}
The $K\times S$ background mean features (or proxies) are kept because the backgrounds among support images are also likely to be inhomogeneous, which gives $\mathbf{u}_n,\;n=1,2,\dots,KS$. 

For the \textbf{feature extraction}, $K$ support samples are individually processed and the prompt synchronization is applied to the $K+1$ prompts, which is formulated by:
\begin{align}
[\mathbf{X}_q^l,\mathbf{P}_q^l] &= B_l([\mathbf{X}_q^{l-1},\mathbf{P}^{l-1}]),
\\
[\mathbf{X}_s^{l,(k)},\mathbf{P}_s^{l,(k)}] &= B_l([\mathbf{X}_s^{l-1,(k)},\mathbf{P}^{l-1}]), \quad k=1,2,\dots,K
\\
\mathbf{P}^l &= \frac1{K+1}\left(\mathbf{P}_q^l + \sum_{k=1}^K\mathbf{P}_s^{l,(k)}\right), 
\end{align}
where $l=1,2, \cdots, L$ enumerates all the transformer blocks.

\section{Detailed Experimeantal Settings}

\subsection{Datasets}
\paragraph{PASCAL-5$^i$} is built from PASCAL VOC 2012~\cite{pascal-voc-2012}~(See the \href{http://host.robots.ox.ac.uk/pascal/VOC/voc2012/}{website} for details.) and SBD~\cite{hariharanSimultaneous2014}(See the \href{http://home.bharathh.info/pubs/codes/SBD/download.html}{website} for details.). We make the dataset splits following~\cite{tianPrior2020}, as shown in Table~\ref{tab:pascal_splits}.

\begin{table}[!thp]
    \caption{Detailed splits of PASCAL-5$^i$}
    \label{tab:pascal_splits}
    \centering
    \begin{tabular}{cl}
        \toprule
         Split & Test classes \\
         \midrule
         PASCAL-5$^0$ & aeroplane, bicycle, bird, boat, bottle  \\
         PASCAL-5$^1$ & bus, car, cat, chair, cow  \\
         PASCAL-5$^2$ & diningtable, dog, horse, motorbike, person  \\
         PASCAL-5$^3$ & potted plant, sheep, sofa, train, tv/monitor \\
         \bottomrule
    \end{tabular}
\end{table}

\paragraph{COCO-20$^i$} is built from COCO 2014~\cite{linMicrosoft2014}~(Licenses of all the images are contained in the annotation file. See the \href{https://cocodataset.org/}{website} for details.). We make the dataset splits following~\cite{tianPrior2020}, as shown in Table~\ref{tab:coco_splits}.
\begin{table}[!thp]
    \caption{Detailed splits of COCO-20$^i$}
    \label{tab:coco_splits}
    \centering
    \begin{tabular}{cl}
        \toprule
         Split & Test classes \\
         \midrule
         COCO-20$^0$ & \makecell{Person, Airplane, Boat, Park meter, Dog, Elephant, Backpack, Suitcase, \\ Sports ball, Skateboard, W. glass, Spoon, Sandwich, Hot dog, Chair, \\ D. table, Mouse, Microwave, Fridge, Scissors}  \\ \midrule
         COCO-20$^1$ & \makecell{Bicycle, Bus, T.light, Bench, Horse, Bear, Umbrella, \\ Frisbee, Kite, Surfboard, Cup, Bowl, Orange, Pizza, \\ Couch, Toilet, Remote, Oven, Book, Teddy} \\ \midrule
         COCO-20$^2$ & \makecell{Car, Train, Fire H., Bird, Sheep, Zebra, Handbag, \\ Skis, B. bat, T. racket, Fork, Banana, Broccoli, Donut, \\ P. plant, TV, Keyboard, Toaster, Clock, Hairdrier} \\ \midrule
         COCO-20$^3$ & \makecell{Motorcycle, Truck, Stop, Cat, Cow, Giraffe, Tie, \\ Snowboard, B. glove, Bottle, Knife, Apple, Carrot, Cake, \\ Bed, Laptop, Cellphone, Sink, Vase, Toothbrush} \\
         \bottomrule
    \end{tabular}
\end{table}

\paragraph{COCO-20$^i\rightarrow$ PASCAL-5$^i$} For the domain shift setting, we make the dataset splits following~\cite{boudiafFewShot2021}, as shown in Table~\ref{tab:coco2pascal}.

\begin{table}[!thp]
    \caption{Detailed splits of COCO-20$^i\rightarrow$ PASCAL-5$^i$}
    \label{tab:coco2pascal}
    \centering
    \begin{tabular}{cl}
        \toprule
         Split & PASCAL-5$^i$ Test classes \\
         \midrule
         COCO-20$^0$ & Airplane, Boat, Chair, D. table, Dog, Person \\ \midrule
         COCO-20$^1$ & Horse, Sofa, Bicycle, Bus \\ \midrule
         COCO-20$^2$ & Bird, Car, P.plant, Sheep, Train, TV \\ \midrule
         COCO-20$^3$ & Bottle, Cat, Cow, Motorcycle \\
         \bottomrule
    \end{tabular}
\end{table}

The construction method of PASCAL-5$^i$ and COCO-20$^i$ follows previous work~\cite{tianPrior2020}. We do not find personally identifiable
information or offensive content in the two datasets.

\subsection{Implementation Details}
We implement FPTrans on vision transformer backbones (\ie, ViT~\cite{dosovitskiyImage2021}, DeiT~\cite{touvronTraining2021}) and a proxy-based classification head. Our experiments are based on ViT-B/16 and DeiT-B/16 (both are pretrained on ImageNet-1K~\cite{dengImageNet2009} with size 224 and finetuned with size 384). We also report the results of two smaller variants, DeiT-S/16 and DeiT-T/16 (pretrained with size 224) in Fig.~\ref{fig:miou}. To enlarge the size of feature maps, we append a residual upsampling layer after the backbone in the baseline and FPTrans. Specifically, given the output of the final transformer block, we reshape it back to 2D feature maps $\mathbf{X}_L\in\mathbb{R}^{C\times H\times W}$, and upsample the feature maps by:
\begin{equation}
    \mathbf{X}_L' = \textrm{Resize}(\mathbf{X}_L) + g(\mathbf{X}_L),
\end{equation}
where Resize is a linear interpolation operation, and $g(\cdot)$ is a bottleneck layer implemented by ``Conv-ReLU-DeConv-ReLU-Conv'', where ``Conv'' is the $1\times 1$ convolutional layer and ``DeConv'' is the $2\times 2$ deconvolutional layer. The hidden channels are set as 256 by default. We observe that a strong pairwise loss~(\ie, a large $\lambda$) leads to over-penalization. Therefore, we experimentally set $\lambda$ to 2e-2 for PASCAL-5$^i$ and smaller 1e-4 for a larger dataset COCO-20$^i$.

\subsection{Algorithm}
The pseudo-code of FPTrans is presented as below. The PyTorch~\cite{paszke2017automatic} and PaddlePaddle\footnote[2]{\url{https://github.com/PaddlePaddle/Paddle}} implementation will be publicly available.

\begin{algorithm}
\caption{Algorithm of FPTrans (Single training step, 1-shot setting)}\label{alg:cap}
\begin{algorithmic}[1]
\Require A training episodes $(I_s, Y_s, I_q, Y_q)$ 
\State $\{\mathbf{a}_{q,p}\}_{p=1}^N, \{\mathbf{a}_{s,p}\}_{p=1}^N \leftarrow I_q, I_s$ \Comment{Reshape images into image patches} 
\State $\mathbf{X}_q\leftarrow\mathtt{Embed}(\{\mathbf{a}_{q,p}\}_{p=1}^N), \mathbf{X}_s\leftarrow\mathtt{Embed}(\{\mathbf{a}_{s,p}\}_{p=1}^N)$ \Comment{Project image patches into embedding}
\State $[\mathbf{x}_{cls}, \mathbf{X}_q] \leftarrow \mathbf{X}_q$, $[\mathbf{x}_{cls}, \mathbf{X}_s] \leftarrow \mathbf{X}_s$ \Comment{Concatenate the class token}
\State $[\mathbf{x}^0, \mathbf{X}_q^0]\leftarrow [\mathbf{x}_{cls}, \mathbf{X}_q] + \mathbf{E}_{pos}$, $[\mathbf{x}^0, \mathbf{X}_s^0]\leftarrow [\mathbf{x}_{cls}, \mathbf{X}_s] + \mathbf{E}_{pos}$ \Comment{Add the positional embedding}
\State $\mathbf{P}^0\leftarrow$ \Call{PromptGeneration}{$[\mathbf{x}_{cls}, \mathbf{X}_s]$} \Comment{Prompt generation}

\State $l=1$
\While{$l \leq L$}
\State $[\mathbf{x}_q^l, \mathbf{X}_q^l, \mathbf{P}_q^l] \leftarrow B_l([\mathbf{x}^{l-1}, \mathbf{X}_q^{l-1}, \mathbf{P}^{l-1}])$
\State $[\mathbf{x}_s^l, \mathbf{X}_s^l, \mathbf{P}_s^l] \leftarrow B_l([\mathbf{x}^{l-1}, \mathbf{X}_s^{l-1}, \mathbf{P}^{l-1}])$
\State $\mathbf{x}^l\leftarrow (\mathbf{x}_q^l + \mathbf{x}_s^l) / 2$
\State $\mathbf{P}^l\leftarrow (\mathbf{P}_q^l + \mathbf{P}_s^l) / 2$
\State $l \leftarrow l + 1$
\EndWhile

\State $\mathbf{F}_q\leftarrow \mathtt{Resize}(\mathbf{X}_q^L) + g(\mathbf{X}_q^L), \mathbf{F}_s\leftarrow \mathtt{Resize}(\mathbf{X}_s^L) + g(\mathbf{X}_s^L)$ \Comment{Upsample features}
\State $[\mathbf{p}_f^L, \{\mathbf{p}_n^L\}_{n=1}^S]\leftarrow  \mathbf{P}^L + g(\mathbf{P}^L)$ \Comment{Project prompt states into the feature space}
\State $\mathbf{u} := [\mathbf{u}_f, \{\mathbf{u}_n\}_{n=1}^S]\leftarrow$ \Call{ProxyGeneration}{$\mathbf{F}_s, Y_s$} \Comment{Feature-based proxy generation}
\State $\mathbf{v} := [\mathbf{v}_f, \{\mathbf{v}_n\}_{n=1}^S]\leftarrow [\frac1G\sum_{j=1}^G\mathbf{p}_{f,j}^L, \{\frac1G\sum_{j=1}^G\mathbf{p}_{n,j}^L\}_{n=1}^S]$ \Comment{Prompt-based proxy generation}
\Statex
\State $\mathcal{P}(\mathbf{F}_{q,i},\mathbf{u}) \leftarrow \frac{\exp({\texttt{sim}(\mathbf{F}_{q,i}, \mathbf{u}_f})/\tau)}
    {
    \exp{(\texttt{sim}(\mathbf{F}_{q,i}, \mathbf{u}_f)/\tau)}+
    \max_n(\exp{(\texttt{sim}(\mathbf{F}_{q,i}, \mathbf{u}_n/\tau)}))
    }$ \Comment{Compute probability}
    
\Statex
\State $\mathcal{L}_{ce}\leftarrow-\sum_{i=1}^{H\times W}(Y_{q,i}\log\mathcal{P}(\mathbf{F}_{q,i},\mathbf{u}) + (1 - Y_{q,i})\log(1 - \mathcal{P}(\mathbf{F}_{q,i},\mathbf{u})))$ 
\Statex \Comment{Feature-proxy based classification loss}

\Statex
\State $\mathcal{L}_{ce}'\leftarrow-\sum_{i=1}^{H\times W}(Y_{q,i}\log\mathcal{P}(\mathbf{F}_{q,i},\mathbf{v}) + (1 - Y_{q,i})\log(1 - \mathcal{P}(\mathbf{F}_{q,i},\mathbf{v})))$ 
\Statex \Comment{Prompt-proxy based Classification loss}

\Statex
\State $\mathcal{L}_{pair} = \frac1{Z}\sum_{(Y_{q,i}+Y_{s,j})\geq 1}\text{BCE}(\sigma(\texttt{sim}(\mathbf{F}_{q,i}, \mathbf{F}_{s,j})/\tau), \mathbf{1}[Y_{q,i}=Y_{s,j}]),$ \Comment{Pairwise loss}

\Statex
\State $\mathcal{L}\leftarrow\mathcal{L}_{ce}+\mathcal{L}_{ce}' + \lambda\mathcal{L}_{pair}$
\end{algorithmic}
\end{algorithm}

\section{More Experimental Results}

\subsection{Detailed Main Results}

See Table~\ref{tab:pascal_supp} and Table~\ref{tab:coco_supp}.

\subsection{Ablation Studies}

\begin{figure}[!thp]
    \centering
    \includegraphics[width=0.6\linewidth]{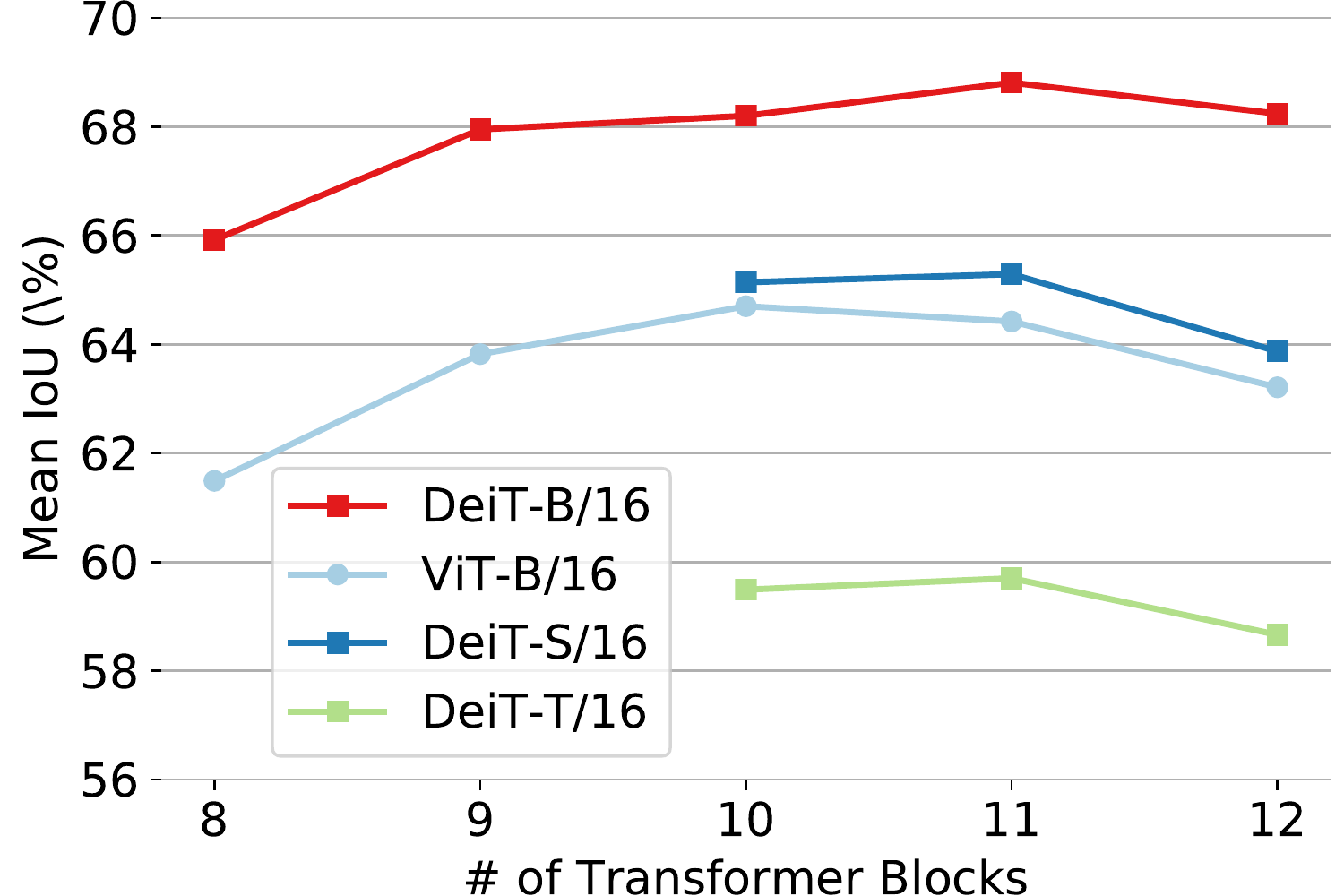}
    \caption{Results with different levels of feature and various backbones on the 1-shot setting. Experiments are conducted on PASCAL-5$^i$.}
    \label{fig:miou}
\end{figure}

\paragraph{Ablations on the transformer blocks. } Previous methods~\cite{Wang_2019_ICCV,Zhang_2019_CVPR,tianPrior2020} use mid-level features because high-level features are prone to lack of details. We inspect different feature levels on various transformer backbones as shown in Fig.~\ref{fig:miou}. 
We observe that FPTrans prefers mid-level features to top-level features, which is consistent with that of CNN-based methods~\cite{Zhang_2019_CVPR,tianPrior2020}. For example,  ViT-B/16 achieves the best results with 10 transformer blocks, while the three DeiT variants (DeiT-B/16, DeiT-S/16, and DeiT-T/16) achieve the best results with 11 blocks. Moreover, The smaller DeiT-S/16 backbone even outperforms the ViT-B/16 backbone, and the smallest DeiT-T/16 backbone with 11 blocks achieves 59.70\% mIoU on par with some ResNet-101 methods, \eg, RePRI~\cite{boudiafFewShot2021} (59.4\%) and PFENet~(60.1\%). For COCO-20$^i$, we find that DeiT-B/16 with 12 transformer blocks gives the best performance. 

\paragraph{Ablations on prompt augmentation.} 
The ablation studies of prompts are listed in Table~\ref{tab:prompt_aug}. From the results in the table, we have two observations.
\begin{itemize}
    \item \#2 (using only the learnable prompts) and \#3 (using only the extracted prompts) actually decreases and increases the accuracy over \#1 (baseline), respectively.
    \item Comparing \#4 against \#3, we observe that adding the learnable prompts brings further improvement.
\end{itemize}
Therefore, in our manuscript, we consider the learnable prompts as a prompt augmentation approach (which should not be used alone).

% We investigate the influence of using prompt augmentation. The results in Table~\ref{tab:prompt_aug} show that the prompt augmentation strategy brings +0.4\% mean IoU gains for PASCAL-5$^i$ and +0.6\% for COCO-20$^i$. This demonstrates the importance of learning discriminative prompts in FPTrans.

\begin{table}[!thp]
    \caption{Ablation studies of the prompting strategy. "Extracted prompts" indicate the foreground and background mean vectors extracted in the prompt generation step. "Learned prompts" indicate the extra added learnable vectors. We conduct experiments with ViT-B/16 on the 1-shot setting.}
    \label{tab:prompt_aug}
    \centering
    \begin{tabular}{ccccc}
        \toprule
         \# &Extracted prompts & Learnable prompts & PASCAL-5$^i$ & COCO-20$^i$ \\
         \midrule
          1 & & & 64.0 & 40.3 \\
          2 & & $\checkmark$ & 63.5 & 39.2 \\
          3 & $\checkmark$ & & 64.3 & 41.4 \\
          4 & \checkmark & $\checkmark$ & \bd64.7 & \bd42.0 \\ 
         \bottomrule
    \end{tabular}
\end{table}

\paragraph{Ablations on the pairwise loss.} 
We recall that the pairwise loss for training FPTrans does NOT pull close background features. The motivation for this design is that we consider the background is not homogeneous. We conduct experiments to show  that enforcing within-class compactness on background features actually compromises FPTrans, as shown in Table~\ref{tab:pairloss}. It is observed that adding the pulling-close to the background features decreases the segmentation accuracy (-0.7\% IoU with 50\% background pairs, and -0.9\% mIoU with all the background pairs). 

\begin{table}[!thp]
    \caption{Results of using the background-background pairs in the pairwise loss. We conduct experiments with ViT-B/16 on the 1-shot setting, and report results on PASCAL-5$^i$.}
    \label{tab:pairloss}
    \centering
    \begin{tabular}{cc}
        \toprule
         Using background-background pairs (percent) & Mean IoU \\
         \midrule
         100\% & 63.8 \\
         50\%  & 64.0 \\
         0\%   & \bd64.7 \\ 
         \bottomrule
    \end{tabular}
\end{table}

\subsection{Qualitative Results}

\begin{figure}
    \centering
    \includegraphics[width=0.9\linewidth]{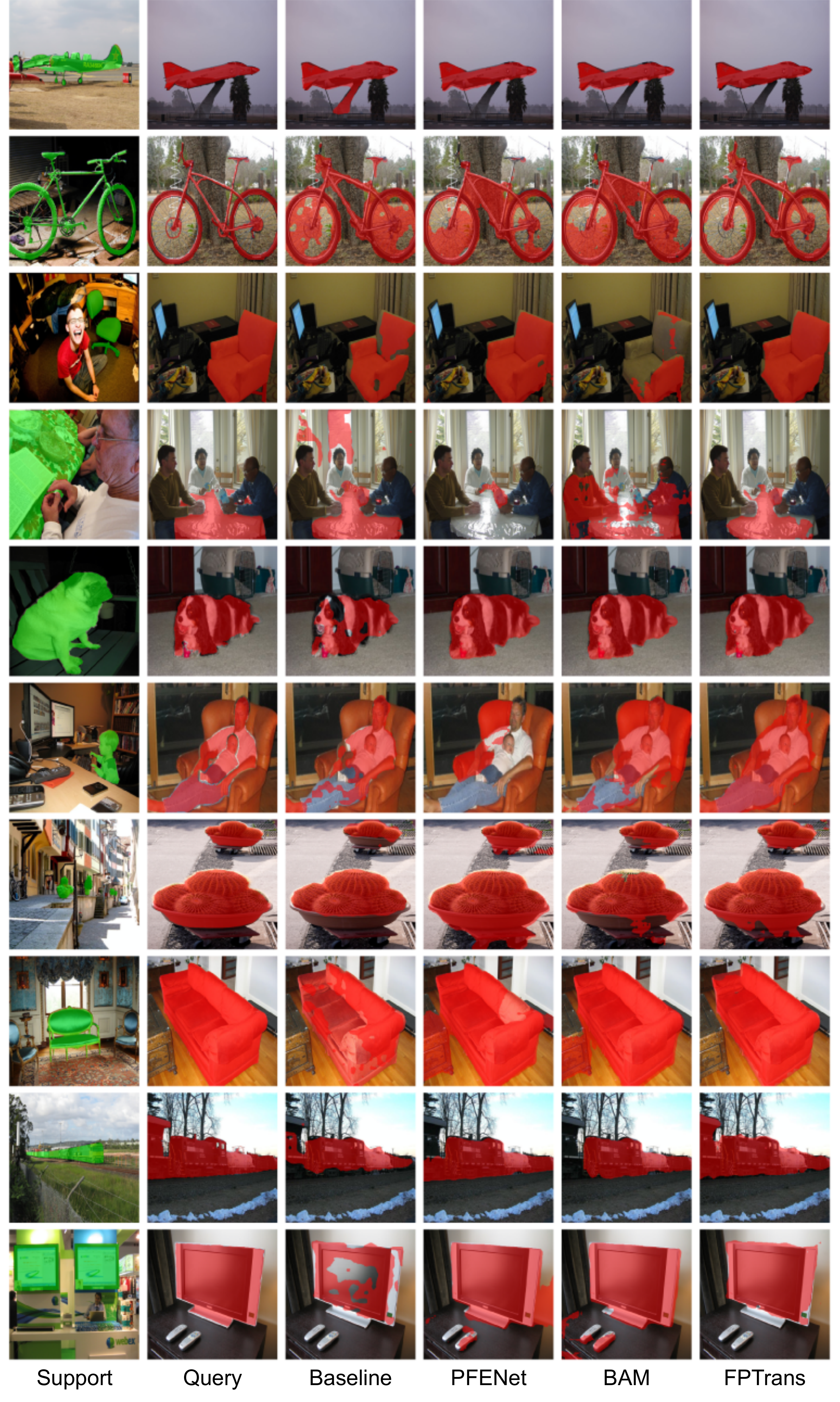}
    \caption{Qualitative comparisons of FPTrans with the baseline and previous methods, PFENet~\cite{tianPrior2020} and BAM~\cite{langLearning2022}.}
    \label{fig:display}
\end{figure}

As presented in Fig.~\ref{fig:display}, we display some qualitative comparisons of FPTrans with the baseline and two previous methods PFENet~\cite{tianPrior2020} and BAM~\cite{langLearning2022}.

\section{Analysis on Computational Cost}
Although FPTrans achieves promising results with a simple framework, it relies on a pretrained model for generating prompts, which consumes extra computational resources. However, the computational cost for generating the prompts is comparable compared with previous methods. 

We add the comparison of the parameter size and FLOPs in Table~\ref{tab:computation_cost}. For a fair comparison, we fix the input size as $480\times480$. We observe that the proposed {\bf FPTrans is actually relatively efficient}, considering its superiority in FSS accuracy. For example, FPTrans with the DeiT-S/16 backbone has 41 Mb parameters and only 80.7 GFLOPs. It is faster than all the competing CNN methods and yet achieves competitive accuracy. Moreover, FPTrans with DeiT-B/16 backbone is superior to the SOTA method BAM w.r.t. both the accuracy (mIoU) and speed (FLOPs).

\begin{table}[!ht]
    \centering
    \caption{Comparison of FPTrans with SOTA methods on the number of parameters and computation cost. $^*$ denotes that the mean IoU is from our experiments.}
    \label{tab:computation_cost}
    \begin{tabular}{ccccc}
    \toprule
    Backbone & Method & Params (M) & GFLOPs & Mean-IoU (\%) \\
    \midrule
    \multirow{4}{*}{ResNet-50}  & PFENet~\cite{tianPrior2020}    & 34    & 231.2 & 60.8 \\
                                & CyCTR~\cite{zhang2021few}     & 37    & 244.7 & 64.2 \\
                                & HSNet~\cite{minHypercorrelation2021}     & 28    & 95.9  & 64.0 \\
                                & BAM~\cite{langLearning2022}       & 52    & 302.2 & 67.8 \\
    \midrule
    \multirow{4}{*}{ResNet-101} & PFENet~\cite{tianPrior2020}    & 53    & 367.9 & 60.1 \\
                                & CyCTR~\cite{zhang2021few}     & 59    & 381.1 & 64.3 \\
                                & HSNet~\cite{minHypercorrelation2021}     & 47    & 145.0 & 66.2 \\
                                & BAM*      & 71    & 438.9 & 67.5 \\
    \midrule
    ViT-B/16    & \multirow{4}{*}{FPTrans}  & 145   & 247.2 & 64.7 \\
    DeiT-T/16   &                           & 11    & 26.7  & 59.7 \\
    DeiT-S/16   &                           & 41    & 80.7  & 65.3 \\
    DeiT-B/16   &                           & 159   & 271.8 & 68.8 \\
    \bottomrule    
    \end{tabular}
\end{table}

Moreover, the computational cost for generating the prompts has the potential to be reduced, because we find that FPTrans is robust to the model for extracting the prompts to some extent. Specifically, when replacing the pretrained ViT with the trained backbone of FPTrans (on PASCAL-5$^i$), the generated prompts can still produce almost the same results. In this way, we only need to save a copy of FPTrans, instead of saving both the pretrained ViT and FPTrans for inference. In future works, we will seek more lightweight prompt-generating models to further reduce the computational cost. 
Since FPTrans only consists of a transformer backbone and a linear classification head, the quantitative computational cost of FPTrans can be directly referred to as vision transformer~\cite{dosovitskiyImage2021, touvronTraining2021}.

\newpage

\begin{sidewaystable}[!thp]
  \caption{Comparison with state-of-the-art methods on PASCAL-5$^i$. We report the mean IoU (\%) and standard deviation averaged on three random seeds under 1-shot and 5-shot settings.}
  \label{tab:pascal_supp}
  \centering
  \small
  \begin{tabular}{clcccccccccc}
    \toprule
    \multirow{2}{*}{Backbone} & \multirow{2}{*}{Method} &  \multicolumn{5}{c}{1-shot} & \multicolumn{5}{c}{5-shot} \\
    \cmidrule{3-12}
      &  & S0 & S1 & S2 & S3 & Mean & S0 & S1 & S2 & S3 & Mean \\
    \midrule
    \multirow{6}{*}{Res-50} & RPMM~\cite{yangPrototype2020} & 55.2 & 66.9 & 52.6 & 50.7 & 56.3 & 56.3 & 67.3 & 54.5 & 51.0 & 57.3 \\
    & PPNet~\cite{liuPartAware2020} & 47.8 & 58.8 & 53.8 & 45.6 & 51.5 & 58.4 & 67.8 & 64.9 & 56.7 & 62.0 \\
    & PFENet~\cite{tianPrior2020} & 61.7 & 69.5 & 55.4 & 56.3 & 60.8 & 63.1 & 70.7 & 55.8 & 57.9 & 61.9 \\
    & CyCTR~\cite{zhang2021few} & 67.8 & 72.8 & 58.0 & 58.0 & 64.2 & 71.1 & 73.2 & 60.5 & 57.5 & 65.6 \\
    & HSNet~\cite{minHypercorrelation2021} & 64.3 & 70.7 & 60.3 & 60.5 & 64.0 & 70.3 & 73.2 & 67.4 & 67.1 & 69.5 \\
    & BAM~\cite{langLearning2022} & 69.0 & 73.6 & 67.6 & 61.1 & 67.8 & 70.6 & 75.1 & 70.8 & 67.2 & 70.9 \\
    \midrule
    \multirow{6}{*}{Res-101} & FWB~\cite{Nguyen_2019_ICCV} & 51.3 & 64.5 & 56.7 & 52.2 & 56.2 & 54.9 & 67.4 & 62.2 & 55.3 & 59.9 \\
    & DAN~\cite{wangFewShot2020a} & 54.7 & 68.6 & 57.8 & 51.6 & 58.2 & 57.9 & 69.0 & 60.1 & 54.9 & 60.5 \\
    & RePRI~\cite{boudiafFewShot2021} & 59.6 & 68.6 & 62.2 & 47.2 & 59.4 & 66.2 & 71.4 & 67.0 & 57.7 & 65.6 \\
    & PFENet~\cite{tianPrior2020} & 60.5 & 69.4 & 54.4 & 55.9 & 60.1 & 62.8 & 70.4 & 54.9 & 57.6 & 61.4 \\
    & CyCTR~\cite{zhang2021few} & 69.3 & 72.7 & 56.5 & 58.6 & 64.3 & 73.5 & 74.0 & 58.6 & 60.2 & 66.6 \\
    & HSNet~\cite{minHypercorrelation2021} & 67.3 & 72.3 & 62.0 & 63.1 & 66.2 & 71.8 & 74.4 & 67.0 & 68.3 & 70.4 \\
    \midrule
    \multirow{2}{*}{ViT-B/16} & Baseline  &    62.9{\tiny$\pm0.36$} &    69.1{\tiny$\pm0.06$} &    62.2{\tiny$\pm0.55$} &    53.0{\tiny$\pm0.30$} &    61.8{\tiny$\pm0.07$} &    70.5{\tiny$\pm1.09$} &    76.0{\tiny$\pm0.32$} &    74.2{\tiny$\pm0.46$} &    65.5{\tiny$\pm1.24$} &    71.5{\tiny$\pm0.77$} \\
    & \cellcolor{grey1} \textbf{FPTrans}   & \cellcolor{grey1} \bd67.1{\tiny$\pm0.47$} & \cellcolor{grey1} \bd69.8{\tiny$\pm0.03$} & \cellcolor{grey1} \bd65.6{\tiny$\pm0.14$} & \cellcolor{grey1} \bd56.4{\tiny$\pm0.75$} & \cellcolor{grey1} \bd64.7{\tiny$\pm0.29$} & \cellcolor{grey1} \bd73.5{\tiny$\pm0.12$} & \cellcolor{grey1} \bd75.7{\tiny$\pm0.04$} & \cellcolor{grey1} \bd77.4{\tiny$\pm0.15$} & \cellcolor{grey1} \bd68.3{\tiny$\pm0.19$} & \cellcolor{grey1} \bd73.7{\tiny$\pm0.08$} \\
    \midrule
    \multirow{2}{*}{DeiT-B/16} & Baseline &    68.2{\tiny$\pm0.34$} &    69.4{\tiny$\pm0.38$} &    61.7{\tiny$\pm1.06$} &    60.5{\tiny$\pm0.66$} &    64.9{\tiny$\pm0.11$} &    75.3{\tiny$\pm0.19$} &    78.1{\tiny$\pm0.12$} &    76.1{\tiny$\pm0.55$} &    73.7{\tiny$\pm0.76$} &    75.8{\tiny$\pm0.11$} \\
    & \cellcolor{grey1} \textbf{FPTrans}   & \cellcolor{grey1} \bd72.3{\tiny$\pm0.49$} & \cellcolor{grey1} \bd70.6{\tiny$\pm0.34$} & \cellcolor{grey1} \bd68.3{\tiny$\pm0.47$} & \cellcolor{grey1} \bd64.1{\tiny$\pm0.49$} & \cellcolor{grey1} \bd68.8{\tiny$\pm0.33$} & \cellcolor{grey1} \bd76.7{\tiny$\pm0.36$} & \cellcolor{grey1} \bd79.0{\tiny$\pm0.03$} & \cellcolor{grey1} \bd81.0{\tiny$\pm0.89$} & \cellcolor{grey1} \bd75.1{\tiny$\pm0.31$} & \cellcolor{grey1} \bd78.0{\tiny$\pm0.14$} \\
    \bottomrule
  \end{tabular}
\end{sidewaystable}

\begin{sidewaystable}[!thp]
  \caption{Comparison with state-of-the-art methods on COCO-20$^i$. We report the mean IoU (\%) and standard deviation averaged on three random seeds under 1-shot and 5-shot settings.}
  \label{tab:coco_supp}
  \centering
  \small
  \begin{tabular}{clcccccccccc}
    \toprule
    \multirow{2}{*}{Backbone} & \multirow{2}{*}{Method} & \multicolumn{5}{c}{1-shot} & \multicolumn{5}{c}{5-shot} \\
    \cmidrule{3-12}
      &  & S0 & S1 & S2 & S3 & Mean & S0 & S1 & S2 & S3 & Mean \\
    \midrule
    \multirow{3}{*}{Res-50} & RePRI~\cite{boudiafFewShot2021} & 32.0 & 38.7 & 32.7 & 33.1 & 34.1 & 39.3 & 45.4 & 39.7 & 41.8 & 41.6 \\
    & HSNet~\cite{minHypercorrelation2021} & 36.3 & 43.1 & 38.7 & 38.7 & 39.2 & 43.3 & 51.3 & 48.2 & 45.0 & 46.9 \\
    & BAM~\cite{wangLearning2022} & 43.4 & 50.6 & 47.5 & 43.4 & 46.2 & 49.3 & 54.2 & 51.6 & 49.6 & 51.2 \\
    \midrule
    \multirow{3}{*}{Res-101} & DAN~\cite{wangFewShot2020a} & - & - & - & - & 24.4 & - & - & - & - & 29.6 \\
    & PFENet~\cite{tianPrior2020} & 34.3 & 33.0 & 32.3 & 30.1 & 32.4 & 38.5 & 38.6 & 38.2 & 34.3 & 37.4 \\
    & HSNet~\cite{minHypercorrelation2021} & 37.2 & 44.1 & 42.4 & 41.3 & 41.2 & 45.9 & 53.0 & 51.8 & 47.1 & 49.5 \\
    \midrule
    \multirow{2}{*}{ViT-B/16} & Baseline  
& 37.3{\tiny$\pm0.31$} 
& 39.6{\tiny$\pm0.20$} 
& 41.5{\tiny$\pm0.30$} 
& 35.3{\tiny$\pm0.35$} 
& 38.4{\tiny$\pm0.15$} 
& 48.2{\tiny$\pm0.36$} 
& 53.5{\tiny$\pm0.16$} 
& 52.9{\tiny$\pm0.16$} 
& 48.8{\tiny$\pm0.15$} 
& 50.8{\tiny$\pm0.20$} \\
    & \cellcolor{grey1} \textbf{FPTrans}   
& \cellcolor{grey1} \bd39.7{\tiny$\pm0.24$} 
& \cellcolor{grey1} \bd44.1{\tiny$\pm0.17$} 
& \cellcolor{grey1} \bd44.4{\tiny$\pm0.23$} 
& \cellcolor{grey1} \bd39.7{\tiny$\pm0.30$} 
& \cellcolor{grey1} \bd42.0{\tiny$\pm0.06$} 
& \cellcolor{grey1} \bd49.9{\tiny$\pm0.29$} 
& \cellcolor{grey1} \bd56.5{\tiny$\pm0.05$} 
& \cellcolor{grey1} \bd55.4{\tiny$\pm0.08$} 
& \cellcolor{grey1} \bd53.2{\tiny$\pm0.34$} 
& \cellcolor{grey1} \bd53.8{\tiny$\pm0.13$} \\
    \midrule
    \multirow{2}{*}{DeiT-B/16} & Baseline 
& 41.8{\tiny$\pm0.55$} 
& 45.4{\tiny$\pm0.75$} 
& 48.8{\tiny$\pm0.19$} 
& 40.3{\tiny$\pm0.58$} 
& 44.1{\tiny$\pm0.47$} 
& 53.9{\tiny$\pm0.53$} 
& 60.1{\tiny$\pm0.75$} 
& 58.9{\tiny$\pm0.55$} 
& 54.4{\tiny$\pm0.80$} 
& 56.8{\tiny$\pm0.64$} \\
    & \cellcolor{grey1} \textbf{FPTrans}
& \cellcolor{grey1} \bd44.4{\tiny$\pm0.27$} 
& \cellcolor{grey1} \bd48.9{\tiny$\pm0.37$} 
& \cellcolor{grey1} \bd50.6{\tiny$\pm0.16$} 
& \cellcolor{grey1} \bd44.0{\tiny$\pm0.36$} 
& \cellcolor{grey1} \bd47.0{\tiny$\pm0.11$} 
& \cellcolor{grey1} \bd54.2{\tiny$\pm0.36$} 
& \cellcolor{grey1} \bd62.5{\tiny$\pm0.20$} 
& \cellcolor{grey1} \bd61.3{\tiny$\pm0.20$} 
& \cellcolor{grey1} \bd57.6{\tiny$\pm0.27$} 
& \cellcolor{grey1} \bd58.9{\tiny$\pm0.06$} \\
    \bottomrule
  \end{tabular}
\end{sidewaystable}

%%%%%%%%%%%%%%%%%%%%%%%%%%%%%%%%%%%%%%%%%%%%%%%%%%%%%%%%%%%%

\end{document}